\PassOptionsToPackage{numbers}{natbib}
\documentclass{article}

\usepackage[main, final]{neurips_2026}
\usepackage{amsmath}
\usepackage{algorithm}
\usepackage[noend]{algpseudocode}

\usepackage[utf8]{inputenc} 
\usepackage[T1]{fontenc}    
\usepackage{hyperref}       
\usepackage{url}            
\usepackage{booktabs}       
\usepackage{amsfonts}       
\usepackage{nicefrac}       
\usepackage{microtype}      
\usepackage{xcolor}         
\usepackage{multirow}

\usepackage{tikz}
\usetikzlibrary{positioning, arrows.meta, calc, fit, backgrounds}
\usepackage{subcaption}
\usepackage{xcolor}
\usepackage{amsmath}

\definecolor{sadColor}{HTML}{DD8452}          
\definecolor{sadOracleColor}{HTML}{CCB974}    

\definecolor{blindColor}{HTML}{4C72B0}        
\definecolor{blindOracleColor}{HTML}{64B5CD}  

\definecolor{exedecColor}{HTML}{55A868}       
\definecolor{gtOracleColor}{HTML}{937860}     

\definecolor{specColor}{HTML}{4D4D4D}
\colorlet{specfill}{specColor!8!white}
\colorlet{subtaskbad}{blindColor!15!white}
\colorlet{subtaskgood}{sadColor!15!white}

\usepackage{acronym}
\acrodef{pbe}[PBE]{Programming-by-Example}
\acrodef{io}[I/O]{input-output}
\acrodef{dsl}[DSL]{Domain-Specific Language}
\acrodef{ai}[AI]{Artificial Intelligence}
\acrodef{agi}[AGI]{Artificial General Intelligence}
\acrodef{arc}[ARC-AGI]{Abstraction and Reasoning Challenge}
\acrodef{llm}[LLM]{Large Language Model}
\acrodef{arc}[ARC-AGI]{Abstraction and Reasoning}
\acrodef{loo}[LOO]{leave-one-out}
\acrodef{gt}[GT]{ground-truth}
\acrodef{sad}[SAD]{Solver-Aware Decomposition}
\acrodef{scst}[SCST]{Self-Critical Sequence Training}
\acrodef{rl}[RL]{Reinforcement Learning}
\acrodef{ce}[CE]{cross-entropy}
\acrodef{sft}[SFT]{supervised fine-tuning}

\title{Solver-Aware Decompositions for Programming-by-Example: When Dividing Requires Knowing How to Conquer}

\author{%
Janis Zenkner \quad Tobias Sesterhenn \quad Tim Grams \quad Christian Bartelt \vspace{2pt} \\ 
Clausthal University of Technology \\
Clausthal-Zellerfeld, Germany \\
\texttt{janis.zenkner@tu-clausthal.de}
}
\begin{document}

\maketitle

\begin{abstract}
Decomposition-based Programming-by-Example splits a task into subtasks: a decomposer predicts intermediate subgoals, and a learned synthesizer generates a program for each subgoal.
Execution-decomposition approaches train the decomposer to imitate \ac{gt} subgoals and thereby treat decomposition quality as a property of the task alone.
A learned synthesizer, however, has its own inductive biases, and \ac{gt} subgoals reflect how the annotator factorized the task, not what this synthesizer can solve.
A decomposer that imitates them can therefore propose subgoals that are logically valid but intractable for the synthesizer.
Motivated by this observation, we propose \ac{sad}, which keeps supervised training on \ac{gt} subgoals as a structural scaffold and additionally trains the decomposer online with self-critical policy gradients against a frozen synthesizer.
Each sampled subgoal is rewarded with the negative cross-entropy of the synthesizer on the target program, a continuous signal of how difficult the subtask is for this synthesizer.
Our experiments show an \emph{accuracy paradox}: with the synthesizer fixed, a \ac{gt}-supervised decomposer agrees more often with the \ac{gt} subgoals, yet its subgoals lead to lower synthesis accuracy, although the synthesizer was trained on the same \ac{gt} data.
On two list-processing domains, \ac{sad} improves synthesis and task accuracy over the \ac{gt}-supervised decomposer, both in-distribution and under length generalization; the same holds under zero-shot transfer to an external benchmark and for a fine-tuned language model.
Even an oracle that receives the \ac{gt} subgoals at inference time, with the same synthesizer and search, fails on tasks that \ac{sad} solves, and fewer than 2\% of \ac{sad}'s solutions to these tasks match the \ac{gt} program.
Decomposition quality is therefore not a property of the task alone: it depends on the synthesizer that executes the subgoals, and the decomposer should be trained against it.\footnote{Code: \url{https://github.com/jzenkner/SAD}}
\end{abstract}
\acresetall

\section{Introduction}
Divide-and-conquer is among the most powerful principles in algorithm design: decompose a complex problem into manageable subproblems, solve each, and recombine~\cite{knuth1998art}.
How useful a given decomposition is, however, depends on the solver that executes it~\cite{allman2018towards}: a partition that a comparison-based sorter can exploit may be of no use to a radix sorter.
The same intermediate structure that one algorithm exploits, another cannot act on at all.
Therefore, a decomposition that is optimal for an ideal solver may be suboptimal for a bounded one, and since practical solvers are bounded~\cite{russell1994provably}, decomposition quality must be defined relative to solver capabilities.

We study this dependence in \ac{pbe}, where a task is specified by \ac{io} examples and the goal is a program in a \ac{dsl} that is consistent with all of them~\citep{gulwani2011automating}.
Consider a task that maps the input list \texttt{[9,1,4,0,1]} to \texttt{[14,4]}.
In a \ac{dsl} with operations such as \texttt{Map}, \texttt{Filter}, and \texttt{ZipWith}, the target program is to subtract 2 from every element, keep the positive values, and double them: \texttt{x1 = Map - 2 x0 | x2 = Filter > 0 x1 | x3 = Map * 2 x2}.
Recent approaches build such programs step by step~\citep{shi2023exedec, zenkner2025transductively}: a decomposer predicts an intermediate subgoal, here the output of the first step \texttt{[7,-1,2,-2,-1]}, and a synthesizer then generates a subprogram that produces it, here \texttt{Map - 2}.
The decomposer thus decides \emph{what} to solve next, and the synthesizer decides \emph{how}.

To learn the first role, execution-decomposition approaches train the decomposer to imitate \ac{gt} subgoals~\citep{shi2023exedec, zenkner2025transductively}, assuming that similarity to \ac{gt} is a reliable proxy for usefulness to the synthesizer.
The assumption is plausible, since the synthesizer is trained on the same \ac{gt} data.
A learned synthesizer, however, develops its own inductive biases and does not reproduce the annotator's choices exactly~\cite{shin2019synthetic}, so a \ac{gt} subgoal can still lie outside what the synthesizer can solve.
Supervised training cannot detect this, because it assigns a similar loss to a subgoal that is close to \ac{gt} but unrealizable and to one that is slightly off but synthesizable.

Take the example from above.
A decomposer trained only on \ac{gt} subgoals predicts \texttt{[7,0,2,3,-1]} as the first subgoal, and since this list matches the \ac{gt} subgoal in three of five positions, the supervised loss is small.
However, no \ac{dsl} operation maps the input to this list (no \texttt{Map} can, because the two \texttt{1}s in the input would have to map to different values), so no subprogram is consistent with this subgoal, and the synthesizer has no correct target to aim for.
Training with synthesizer feedback separates these cases: a subgoal from which the synthesizer cannot produce the target subprogram receives a low reward however close it is to \ac{gt}, while a deviation the synthesizer can act on is penalized less.
A decomposer trained this way therefore predicts the realizable subgoal \texttt{[7,-1,2,-2,-1]}, from which the synthesizer solves the entire task.
Closeness to \ac{gt} did not make the first subgoal useful; what decides is whether the synthesizer can act on it.
Appendix~\ref{app:failure_cases} gives two further cases from our evaluation: a subgoal that no single operation produces (Figure~\ref{fig:struct_mismatch_example}), and a realizable \ac{gt} path that the synthesizer cannot follow (Figure~\ref{fig:proc_mismatch_example}).

\begin{figure}[b]
     \centering
     \begin{subfigure}[h]{0.49\textwidth}
         \centering
         \includegraphics[width=\textwidth]{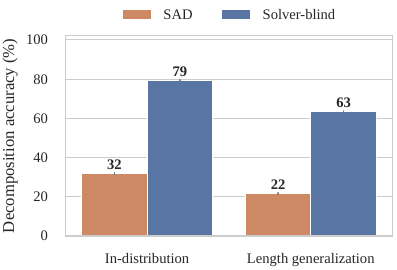}
         \caption{Decomposition accuracy}
         \label{fig:paradox_decomp}
     \end{subfigure}
     \hfill
     \begin{subfigure}[h]{0.49\textwidth}
         \centering
         \includegraphics[width=\textwidth]{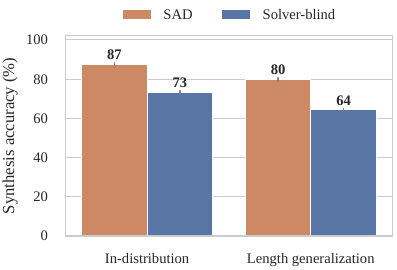}
         \caption{Synthesis accuracy}
         \label{fig:paradox_synth}
     \end{subfigure}
     \caption{Accuracy paradox on Deepcoder. (a) Decomposition accuracy: predicted subgoal matches the \ac{gt} subgoal exactly. (b) Synthesis accuracy: the synthesizer produces the correct subprogram from the predicted subgoal. Both are single-step metrics under teacher forcing. Solver-blind training reaches higher decomposition accuracy but lower synthesis accuracy than solver-aware training.}
     \label{fig:accuracy_paradox}
\end{figure}

The same pattern appears in aggregate.
We call a decomposer that is trained only on \ac{gt} subgoals \emph{solver-blind}.
With the synthesizer fixed, it matches the \ac{gt} subgoals more often than a decomposer trained with synthesizer feedback, yet its subgoals lead to lower synthesis accuracy (Figure~\ref{fig:accuracy_paradox}).
We refer to this result as the \emph{accuracy paradox} (Section~\ref{sec:accuarcy_paradox}): the solver-blind decomposer learns how the annotator decomposes tasks, which is not how the synthesizer solves them most easily.

Accounting for the downstream solver is not new in itself: prior work selects supervision targets with the solver in mind, for example by choosing among \ac{gt}-derived sketches by their estimated enumeration cost~\citep{nye2019learning}.
We instead train the decomposer online against the learned synthesizer that executes its subgoals, which leads to the following contributions:
\begin{enumerate}
    \item We introduce \ac{sad}, which keeps supervised training on \ac{gt} subgoals as a structural scaffold and adds \ac{scst} against a frozen learned synthesizer.
    The synthesizer scores each sampled decomposition online by its \ac{ce} on the \ac{gt} program, a continuous reward that stays informative for partially correct programs.
    Because the greedy decomposition serves as baseline, the advantage measures how much easier the sampled subgoal is for the synthesizer than the greedy one.
    \item On Deepcoder and Lambdabeam, the accuracy paradox holds in every setting, and \ac{sad} improves task accuracy over the solver-blind decomposer both in-distribution and under length generalization (Section~\ref{sec:performance}).
    The improvement carries over to zero-shot transfer to the benchmark of \citet{rule2024symbolic} (Appendix~\ref{app:rule}) and to a fine-tuned Llama-3.1-8B (Section~\ref{sec:llm}).
    On Robustfill, where the decomposition order is fixed, both decomposers perform equally, as predicted (Section~\ref{sec:robustfill}).
    \item With the same synthesizer and search, \ac{sad} solves tasks that an oracle with \ac{gt} subgoals fails, and fewer than 2\% of these solutions match the \ac{gt} program (Section~\ref{sec:gtoracle}).
    \ac{gt} decompositions are therefore not optimal for this synthesizer, and different decomposers reach different programs, which makes decomposition diversity a resource for future work, for example for ensembles or robustness under distribution shift.
\end{enumerate}

\section{Related Work}
\label{sec:relwork}
The central claim of this paper rests on a specific structural condition: a learned decomposer interacting with a bounded, learned synthesizer whose tractable distribution can diverge from the annotator's choices.
This condition is absent in much prior work, and recognizing precisely where it does and does not arise is what defines the scope of \ac{sad}'s contribution.

Enumerative~\citep{solar2008program,gulwani2011automating,alur2013syntax,feser2015synthesizing}, neurally guided~\citep{balog2016deepcoder,yin2017syntactic,lee2018accelerating}, and \ac{llm}-based approaches~\citep{qiu2023phenomenal,li2024programming,piriyakulkij2024doing,olausson2023self,madaan2023self} generate the program from the task specification without a separate decomposer.
In these approaches, there is no decomposer--synthesizer interface to align.

Other approaches use intermediate structure, such as sketches~\citep{murali2017neural}, latent codes~\citep{hong2021latent}, plans or as-needed decompositions~\citep{zhang2023planning,prasad2023adapt,witt2023divide}, and compositional modules~\citep{chen2020compositional,klinger2023compositional}.
Hierarchical programmatic \ac{rl} methods such as LEAPS~\citep{trivedi2021learning} and HPRL~\citep{liu2023hierarchical} execute their programs with a deterministic interpreter.
In both groups, the intermediate structure is either produced and consumed by one model or executed by a fixed procedure, so there is no separately trained learned consumer whose preferences can diverge from the annotation.

REINFORCE~\citep{williams1992simple}, sequence-level training~\citep{ranzato2015sequence}, \ac{scst}~\citep{rennie2017self}, and CodeRL~\citep{le2022coderl} optimize the model that produces the final output.
\ac{sad} uses the same machinery, but the optimized model produces subgoals that a separate, frozen synthesizer consumes.
The feedback therefore changes the problem handed to the synthesizer, not the synthesizer's solution.
Adaptive curriculum methods~\citep{graves2017automated,matiisen2019teacher,portelas2020automatic} also use downstream performance to guide learning, but they select tasks from a fixed distribution instead of generating intermediate problems.

Exedec~\citep{shi2023exedec} and TIIPS~\citep{zenkner2025transductively} use the same pipeline as \ac{sad}: a learned decomposer predicts subgoals, and a learned synthesizer generates subprograms for them.
Both train the decomposer to imitate \ac{gt} subgoals, so the setting above arises, but the synthesizer's behavior is never used as a training signal.
\ac{sad} keeps this pipeline and changes how the decomposer is trained.
SketchAdapt~\citep{nye2019learning} goes beyond \ac{gt} imitation.
It derives candidate sketches from the \ac{gt} program and estimates the enumeration time of each from the likelihood under a learned recognizer.
The sketch generator is then trained to imitate a candidate whose estimate fits a sampled time budget.
SketchAdapt thus accounts for the solver through a proxy for its capabilities and does not run synthesis during training.
\ac{sad} instead samples subgoals from the current policy, scores each sample by the \ac{ce} of the frozen synthesizer on the target program, and updates the policy with \ac{scst}.

Outside program synthesis, solver feedback is used to train feasibility predictors for a fixed task-and-motion planner~\citep{yang2022sequence,wells2019learning}, whereas \ac{sad} updates the model that generates the intermediate representation.
\citet{inala2020synthesizing} adapt supervision through trajectories generated by a teacher.
In HyperTree Proof Search~\citep{lample2022hypertree}, the learned policy is part of the prover and is trained together with proof search, whereas the synthesizer in \ac{sad} is trained separately and then frozen.

In summary, \ac{sad} combines four elements: a frozen learned synthesizer whose predictive distribution defines the reward, a continuous \ac{ce} reward instead of binary success, online scoring of every decomposition sampled by the current policy, and policy-gradient updates of the decomposer.

\section{Background: Decomposition-based Synthesis}
\label{sec:background}
In \ac{pbe}, a task is given as a set of \ac{io} examples $\mathcal{E} = \{(I_i, O_i)\}_{i=1}^n$, and the goal is to find a program $p$ in a \ac{dsl} such that $p(I_i) = O_i$ for all $(I_i, O_i) \in \mathcal{E}$~\cite{gulwani2011automating}.
A \ac{dsl} is a small programming language whose operations are tailored to one class of tasks, such as list or string transformations~\cite{gulwani2017program}.
Restricting programs to such a language keeps the search space small while the task-specific operations keep the target programs expressible, so the design of a \ac{dsl} trades solution coverage against search tractability~\cite{solar2008program, cambronero2023flashfill++}.
We build on Exedec~\cite{shi2023exedec}, which solves a task by decomposing it sequentially into smaller synthesis problems: it alternates decomposition, synthesis, execution, and state update until the specification is satisfied or a step limit $T$ is reached.
\begin{algorithm}[b]
\caption{Inference loop of decomposition-based synthesis (Exedec).}
\label{algo:synth}
\begin{algorithmic}[1]
\Require Decomposition step limit $T$

\Function{GenerateProgram}{$\{(I_i, O_i)\}$}

\State $t \gets 1$
\State $(I_i^{(1)}, O_i^{(1)}) \gets (I_i, O_i), \; \forall i$

\While{$t \leq T$}

    \State $\{O_{\text{pred}, i}^{(t)}\} \gets \text{DecompositionModel}(\{(I_i^{(t)}, O_i^{(t)})\})$
    \Comment{Predict subgoals}

    \State $P^{(t)} \gets \text{Synthesizer}(\{(I_i^{(t)}, O_{\text{pred}, i}^{(t)})\})$
    \Comment{Generate subprogram}

    \State $E_i^{(t)} \gets \text{Execute}(P^{(t)}, I_i^{(t)}), \; \forall i$

    \If{$\forall i.\; E_i^{(t)} = O_i$} \Comment{Solution found}
        \State \Return $\text{Combine}(P^{(1)}, \dots, P^{(t)})$
    \EndIf

    \If{RobustFill} \Comment{Update program state (domain-specific)}
    \vspace{2pt}
        \State $(I_i^{(t+1)}, O_i^{(t+1)}) \gets (I_i^{(t)},\; \text{REMOVEPREFIX}(O_i^{(t)}, E_i^{(t)}))$
    \Else
        \State $(I_i^{(t+1)}, O_i^{(t+1)}) \gets (I_i^{(t)} \cup E_i^{(t)},\; O_i^{(t)})$
    \EndIf

    \State $t \gets t + 1$

\EndWhile

\State \Return \text{Failure}

\EndFunction

\end{algorithmic}
\end{algorithm}

Each iteration $t$ transforms the current state $(I^{(t)}, O^{(t)})$, which initially is the task itself, in four stages.
In the \emph{decomposition} stage, the decomposer reads the current state and predicts an intermediate subgoal $O^{(t)}_{\text{pred}}$, i.e., the outputs that the next subprogram should produce from the current inputs.
Together with these inputs, the subgoal forms the subtask $(I^{(t)}, O^{(t)}_{\text{pred}})$, which is atomic: it is meant to be solved by a single minimal subprogram.
In the \emph{synthesis} stage, the synthesizer, a learned sequence model, reads the \ac{io} examples of this subtask and generates such a subprogram $P^{(t)}$, aiming to map each input $I^{(t)}_i$ to its subgoal $O^{(t)}_{\text{pred},i}$.
In the \emph{execution} stage, $P^{(t)}$ is run on the current inputs, which yields the results $E^{(t)}_i = P^{(t)}(I^{(t)}_i)$.
If these results equal the target outputs $O_i$ for all examples, the subprograms found so far form the solution; otherwise, the \emph{update} stage incorporates $E^{(t)}$ into the state, and the next iteration begins.
The execution results are never compared with the predicted subgoal, so a subprogram whose output differs from $O^{(t)}_{\text{pred}}$ is still kept; the subgoal thus guides the synthesizer without constraining it.
Algorithm~\ref{algo:synth} describes inference, where the synthesizer receives the predicted subtask; during training, the \ac{sad} synthesizer is conditioned on the full current task instead (Section~\ref{sec:synth_training}).

The update stage depends on the domain, and it determines how much freedom the decomposer has.
In Deepcoder and Lambdabeam~\cite{balog2016deepcoder, shi2023lambdabeam}, each step adds its output to the inputs as a new variable, while the target $O^{(t)}$ stays fixed.
A subgoal that is plausible locally but does not lead toward the target therefore passes unnoticed at that step: it uses up one of the $T$ steps, and later decompositions are conditioned on it.
This is why the cost of misaligned subgoals grows with program length (Appendix~\ref{app:perfvslen}).
In Robustfill~\cite{devlin2017robustfill}, by contrast, each step removes its output from the front of the remaining target string, so the residual shows what is left to produce.
Since each step must produce a prefix of this residual, the decomposition order is fixed; Section~\ref{sec:eval} discusses what this means for the evaluation.

If no solution is found within $T$ iterations, the loop returns a failure.
At test time, a beam search runs several rollouts of this loop in parallel (Appendix~\ref{app:beamsearch}).

\section{\acl{sad}}
\label{sec:method}
\ac{sad} keeps the pipeline of Section~\ref{sec:background} and changes only how the decomposer is trained.
A synthesizer is first trained and then frozen (Section~\ref{sec:synth_training}); the decomposer is then trained against it with \ac{gt} supervision and \ac{scst} (Section~\ref{sec:sad_training}), and at inference both models are combined in a beam search (Section~\ref{sec:inference}).
Throughout, we write $s_t = (I^{(t)}_i, O^{(t)}_i)$ for the state at step $t$ and $p^*_t$ and $O^*_t$ for the \ac{gt} subprogram and the \ac{gt} subgoal of this step.

\subsection{Synthesizer Training}
\label{sec:synth_training}
The synthesizer is trained with teacher forcing to generate the next \ac{gt} subprogram from the full current task, by minimizing the \ac{ce} over program tokens,
\[
\mathcal{L}_{\text{syn}} = \mathrm{CE}\big(\mathrm{Synth}(s_t),\, p^*_t\big).
\]
Unlike the synthesizer used in the original Exedec study~\cite{shi2023exedec}, which is trained on \ac{gt} subtasks, it must therefore infer which subtask comes next as well as how to implement it.
This makes the synthesizer informative also on subgoals that leave the \ac{gt} trajectory, which is where \ac{sad} needs its feedback.
At inference, the synthesizer receives the subtask $(I^{(t)}, O^{(t)}_{\text{pred}})$ induced by the predicted subgoal, as in Algorithm~\ref{algo:synth}.
After training, the synthesizer is frozen and serves as the fixed solver against which the decomposer learns to propose subgoals.
The solver-blind baseline uses the same synthesizer (Section~\ref{sec:eval}, Appendix~\ref{app:training}).

\subsection{Solver-Aware Training}
\label{sec:sad_training}
The decomposer defines a distribution $\pi_\theta(O \mid s_t)$ over subgoals $O$.
\ac{sad} rewards a subgoal by how well the frozen synthesizer recovers the \ac{gt} subprogram from the subtask that the subgoal induces:
\[
R(O) = -\mathrm{CE}\big(\mathrm{Synth}(I^{(t)}, O),\, p^*_t\big).
\]
A subgoal thus receives a high reward if it places the synthesizer where $p^*_t$ is likely.
In this reward, $p^*_t$ fixes \emph{what} must be synthesized, and the subgoal determines \emph{how} easy this is for the synthesizer; the \ac{gt} subgoal does not enter it.
The \ac{gt} program serves as a single verifiable target, since accepting alternative correct programs would require deciding semantic equivalence between programs.
Because the \ac{ce} is continuous, the reward also grades subgoals from which the synthesizer generates a partially correct program.

\begin{figure}[t]
\centering
\resizebox{\linewidth}{!}{%
\begin{tikzpicture}[
    box/.style={draw=gray!70, fill=gray!6, rectangle, rounded corners=4pt, align=center,
                minimum height=1.1cm, minimum width=2.4cm, thick},
    frozen/.style={box, fill=gray!28},
    goal/.style={box, minimum width=3.4cm},
    arr/.style={-{Stealth[length=5pt]}, thick},
    lbl/.style={draw=none, fill=none, inner sep=2pt, font=\small}
]

\node[box, fill=sadColor!20, draw=sadColor]
    (decomp) {Decomposer \\ $\pi_\theta$};

\node[goal, above right=0.7cm and 1.4cm of decomp]
    (sub_s) {Sampled subgoal \\ $O^s \sim \pi_\theta(\cdot \mid s_t)$};

\node[goal, below right=0.7cm and 1.4cm of decomp]
    (sub_g) {Greedy subgoal \\ $O^g$};

\node[frozen, right=1.2cm of sub_s]
    (syn_s) {Frozen \\ synthesizer};

\node[frozen, right=1.2cm of sub_g]
    (syn_g) {Frozen \\ synthesizer};

\node[box, right=1.2cm of syn_s]
    (rew_s) {Reward \\ $R(O^s)$};

\node[box, right=1.2cm of syn_g]
    (rew_g) {Reward \\ $R(O^g)$};

\node[box, minimum width=1.6cm,
      right=1.2cm of $(rew_s)!0.5!(rew_g)$]
    (adv) {Advantage \\ $A = R(O^s) - R(O^g)$};

\node[box, below=1.0cm of decomp]
    (gt) {\acs{gt} subgoal \\ $O^*_t$};

\draw[arr] ($(decomp.north) + (0, 1.0)$)
    -- node[lbl, right] {state $s_t$}
    (decomp.north);

\draw[arr] (decomp.north east) -- (sub_s.west);
\draw[arr] (decomp.south east) -- (sub_g.west);
\draw[arr] (sub_s) -- (syn_s);
\draw[arr] (sub_g) -- (syn_g);
\draw[arr] (syn_s) -- node[lbl, above] {$p^s_t$} (rew_s);
\draw[arr] (syn_g) -- node[lbl, below] {$p^g_t$} (rew_g);
\draw[arr] (rew_s.east) -| (adv.north);
\draw[arr] (rew_g.east) -| (adv.south);

\draw[arr, dashed] (adv.west)
    -- node[lbl, below] {$\mathcal{L}_{\text{RL}}$ (\acs{scst})}
    (decomp.east);

\draw[arr, dashed] (gt.north)
    -- node[lbl, left] {$\mathcal{L}_{\text{sup}}$}
    (decomp.south);

\begin{scope}[shift={($(syn_g.south)!0.5!(rew_g.south) + (0,-0.9)$)},
              local bounding box=legend]
    \draw[arr] (-3.0,0) -- ++(0.8,0)
        node[lbl, right] {data flow};
    \draw[arr, dashed] (0.0,0) -- ++(0.8,0)
        node[lbl, right] {training update};
\end{scope}
\draw[gray!50, rounded corners=2pt]
    ($(legend.south west) + (-0.15,-0.15)$) rectangle
    ($(legend.north east) + (0.15,0.15)$);
    
\end{tikzpicture}%
}
\caption{One \ac{sad} training step. For the state $s_t$, the decomposer samples a subgoal $O^s$ and decodes a greedy subgoal $O^g$. The frozen synthesizer scores both by its \ac{ce} on the \ac{gt} subprogram $p^*_t$, and the advantage $A$ weights the policy-gradient update $\mathcal{L}_{\text{RL}}$. The supervised loss $\mathcal{L}_{\text{sup}}$ on the \ac{gt} subgoal $O^*_t$ is applied in the same step.}
\label{fig:sad_pipeline_detailed}
\end{figure}

We optimize $\pi_\theta$ with \ac{scst}~\cite{rennie2017self}; Figure~\ref{fig:sad_pipeline_detailed} shows one training step.
For each training state $s_t$, \ac{sad} proceeds as follows:
\begin{enumerate}
    \item Sampling: draw a subgoal $O^s \sim \pi_\theta(\cdot \mid s_t)$.
    \item Baseline: decode the greedy subgoal $O^g$ from $\pi_\theta(\cdot \mid s_t)$.
    \item Evaluation: compute $R(O^s)$ and $R(O^g)$ with the frozen synthesizer.
    \item Advantage: $A = R(O^s) - R(O^g)$.
    \item Update: increase the log-probability of $O^s$ if $A > 0$ and decrease it if $A < 0$, i.e., minimize
    \[
    \mathcal{L}_{\text{RL}} = -\mathbb{E}_{O^s \sim \pi_\theta(\cdot \mid s_t)}\big[A \cdot \log \pi_\theta(O^s \mid s_t)\big].
    \]
\end{enumerate}
Since both subgoals are scored by the same synthesizer against the same target, $A$ measures how much easier the synthesizer finds the sampled subgoal than the greedy one.
If both subgoals lead to the same program, $A$ compares the synthesizer's confidence in this program under each; if they lead to different programs, $A$ also favors the subgoal under which the synthesizer is closer to $p^*_t$ (Appendix~\ref{app:gtreward}).
\ac{scst} is REINFORCE~\cite{williams1992simple} with the greedy subgoal of the current policy as baseline, which reduces the variance of the gradient without a learned value model.
Each training state yields one subgoal whose reward the frozen synthesizer returns directly, so there is no long-horizon credit assignment that would call for a value-based method such as PPO~\cite{schulman2017proximal}.

The supervised loss $\mathcal{L}_{\text{sup}} = \mathrm{CE}\big(\pi_\theta(\cdot \mid s_t), O^*_t\big)$ on the \ac{gt} subgoal complements $\mathcal{L}_{\text{RL}}$: it keeps the decomposer in the space of meaningful subgoals, and $\mathcal{L}_{\text{RL}}$ moves it within this space toward subgoals that the synthesizer can solve.
Without $\mathcal{L}_{\text{sup}}$, the policy collapses to degenerate subgoals that the synthesizer tolerates on training data, with a training decomposition accuracy of 2--9\% and a test task accuracy below 20\% (single seed, Appendix~\ref{app:lsupablation}).
The full training objective is
\[
\mathcal{L} = \mathcal{L}_{\text{sup}} + \mathcal{L}_{\text{RL}} - \lambda H(\pi_\theta), \quad \lambda = 0.01,
\]
where the entropy term $H(\pi_\theta)$ keeps the sampled subgoals diverse and prevents early convergence to deterministic decompositions.

\subsection{Inference}
\label{sec:inference}
At inference, \ac{sad} runs the loop of Algorithm~\ref{algo:synth}.
Instead of a single rollout, a beam search runs over full decomposition-synthesis trajectories, where each beam holds a partial program.
Candidates are scored by the sum of the decomposer and synthesizer log-probabilities and pruned if they fail to execute or are redundant with a higher-scoring candidate; all methods use a beam size of 10 (Appendix~\ref{app:beamsearch}).

\section{Experiments with Transformers Trained from Scratch}
\label{sec:experiments}
We train Transformer decomposers and synthesizers from scratch on three \ac{pbe} domains and compare \ac{sad} with a solver-blind baseline that shares its synthesizer, data, and search, so that the two differ only in the training signal of the decomposer.
Section~\ref{sec:llm} repeats this comparison with pretrained \acp{llm}.

\subsection{Evaluation}
\label{sec:eval}

\textbf{Domains} \quad
Deepcoder~\citep{balog2016deepcoder} operates on integer lists with a \ac{dsl} of first- and higher-order functions such as \texttt{Map} and \texttt{Filter}.
The output of each step becomes a named variable that later steps can use, and a task usually admits several valid programs and decompositions.
Lambdabeam~\citep{shi2023lambdabeam} extends Deepcoder with lambda functions built during synthesis and a conditional \texttt{If}, which enlarges the search space, so a subgoal that the synthesizer cannot act on costs more search.
Robustfill~\citep{devlin2017robustfill} covers string transformations, where a program is a concatenation of subprograms.
Its residual state update fixes the decomposition order (Section~\ref{sec:background}), so the decomposer has no alternative decompositions to choose among.
We therefore use Robustfill as a negative control: if the gains of \ac{sad} come from adapting decompositions to the synthesizer, they should vanish in this domain.
Appendix~\ref{app:domains} gives the \acp{dsl}, example tasks, and data generation.

\textbf{Settings} \quad
Each domain is evaluated in-distribution on a standard test split and under length generalization, where models are trained on shorter programs and tested on longer ones: lengths 1--4 for training and 5 for testing in the list domains, and lengths 1--6 for training and 7--10 for testing in Robustfill.
In both settings, training and test tasks are generated by the same procedure from the same \ac{dsl}.
To test whether the gains carry over to tasks from an independent source, we additionally use the list-function benchmark of \citet{rule2024symbolic}, which is part of BIG-bench~\citep{srivastava2023beyond}.
Its 250 tasks map integer lists to integer lists and were designed to study how humans learn rules, not generated from a synthesis \ac{dsl}.
We evaluate the Lambdabeam-trained models without retraining on the 129 deduplicated tasks for which enumerative search finds a Lambdabeam program, so this setting is a zero-shot transfer to an external task distribution (Appendix~\ref{app:rule}).

\textbf{Methods} \quad
\ac{sad} differs from Exedec~\citep{shi2023exedec} in two respects: the input on which the synthesizer is trained (Section~\ref{sec:synth_training}) and the training signal of the decomposer.
To separate the two, we compare three methods:
Exedec, with a synthesizer trained on \ac{gt} subtasks and a decomposer trained on \ac{gt} subgoals;
the solver-blind baseline, which is Exedec with the synthesizer of \ac{sad}, trained on the full current task, and a decomposer trained on \ac{gt} subgoals;
and \ac{sad}, with the same synthesizer as the solver-blind baseline and a decomposer trained with $\mathcal{L}_{\text{sup}}$ and \ac{scst}.
At inference, all three methods give the synthesizer the subtask induced by the predicted subgoal.
Since \ac{sad} and the solver-blind baseline share architecture, synthesizer weights, training data, and search, our conclusions rest on this pair; Exedec connects the results to prior work, and its difference to \ac{sad} combines both changes.
Appendix~\ref{app:training} details the training of each method.

\textbf{\ac{gt} oracle} \quad
The \ac{gt} oracle replaces the predicted subgoals with \ac{gt} subgoals at inference and uses the same synthesizer and beam as \ac{sad}.
It gives an approximate ceiling for a decomposer that reproduces \ac{gt} exactly, and the tasks that \ac{sad} solves but the oracle does not show where \ac{gt} subgoals are not the best choice for this synthesizer.
Appendix~\ref{app:beamoracles} adds a beam oracle that separates which decompositions a model generates from how it ranks them.

\textbf{Metrics} \quad
The primary metric is task accuracy, the percentage of tasks for which the synthesized program satisfies all \ac{io} examples, including a held-out one.
Two single-step diagnostics, both measured under teacher forcing, complement it.
Decomposition accuracy is the percentage of steps in which the predicted subgoal equals the \ac{gt} subgoal, and synthesis accuracy is the percentage of steps in which the synthesizer generates exactly the \ac{gt} subprogram from the predicted subgoal.
The first measures agreement with \ac{gt}, the second how useful the subgoals are to the synthesizer; together they expose the accuracy paradox.

\textbf{Setup} \quad
All models are 3-layer Transformers with embedding dimension 512 and hidden dimension 1024.
Inference uses a beam size of 10 and a step limit of $T = 10$ for the list domains and $T = 20$ for Robustfill.
We follow a leave-one-out protocol in which the models receive $n-1$ \ac{io} examples and a solution counts only if it also satisfies the held-out example.
Each result is averaged over 5 seeds with 1000 test tasks per seed, and differences are tested with paired $t$-tests at the 5\% level.
Appendix~\ref{app:train_and_inf} gives the full training and inference details.

\subsection{Results and Discussion}
\label{sec:results}
\ac{sad} solves more tasks than the solver-blind baseline, although its subgoals agree less often with the \ac{gt} subgoals.
Its subgoals also lead to a higher synthesis accuracy, even though the synthesizer was trained on the same \ac{gt} data, and they let \ac{sad} solve tasks on which even an oracle with \ac{gt} subgoals fails.
On Robustfill, where the decomposition order is fixed, the gains disappear, as predicted.

\begin{figure}[t]
     \centering
     \begin{subfigure}[b]{0.48\textwidth}
         \centering
         \includegraphics[width=\textwidth]{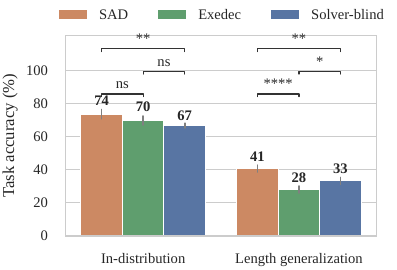}
         \caption{Deepcoder}
         \label{fig:perf_dc}
     \end{subfigure}
     \hfill
     \begin{subfigure}[b]{0.48\textwidth}
         \centering
         \includegraphics[width=\textwidth]{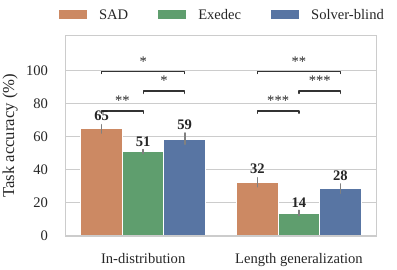}
         \caption{Lambdabeam}
         \label{fig:perf_lb}
     \end{subfigure}
     \caption{Task accuracy on Deepcoder and Lambdabeam, in-distribution and under length generalization, averaged over 5 seeds. Significance levels of the paired $t$-tests: ns: $p>0.05$, *: $p<0.05$, **: $p<0.01$, ***: $p<0.001$, ****: $p<0.0001$.}
     \label{fig:performance_lists}
\end{figure}

\textbf{Solver-aware training solves more tasks} \quad
\label{sec:performance}
Figure~\ref{fig:performance_lists} reports task accuracy on the list domains.
In both domains and both settings, \ac{sad} solves more tasks than the solver-blind baseline.
On Deepcoder, it reaches $73.6\% \pm 3.8\%$ against $66.5\% \pm 1.8\%$ in-distribution and $40.7\% \pm 2.6\%$ against $33.3\% \pm 2.7\%$ under length generalization; on Lambdabeam, it reaches $64.9\% \pm 3.2\%$ against $58.5\% \pm 4.4\%$ and $32.5\% \pm 3.5\%$ against $28.5\% \pm 3.4\%$.
All four differences are statistically significant (paired $t$-tests, Appendix~\ref{app:pvalues}).
The gain also carries over to tasks from an independent source: without retraining, the Lambdabeam-trained \ac{sad} model solves $42.3\% \pm 7.1\%$ of the Rule et al.\ tasks, against $33.0\% \pm 3.5\%$ for the solver-blind baseline (Appendix~\ref{app:rule}).
Since the two methods share synthesizer, data, and search, this shows that the training signal of the decomposer accounts for the gain, and the transfer result indicates that the learned subgoals remain usable on tasks outside the training distribution.
Exedec lies between the two methods on Deepcoder in-distribution and below both in the other three settings.

Compared to the solver-blind baseline, the relative gain is larger under length generalization than in-distribution: $22.4\%$ against $10.7\%$ on Deepcoder and $13.9\%$ against $10.9\%$ on Lambdabeam.
The gap also widens with program length (Appendix~\ref{app:perfvslen}), although both decomposers are trained on single steps only.
This indicates that solver-aware training matters more the further the test tasks lie from the training data.
Two reasons are likely, although we have not tested them separately: on task structures not seen during training, imitating \ac{gt} subgoals offers less guidance, and under the state update of the list domains (Section~\ref{sec:background}), a subgoal that the synthesizer cannot act on uses up a step and conditions all later steps, so errors accumulate with program length.

\textbf{The accuracy paradox} \quad
\label{sec:accuarcy_paradox}
Table~\ref{tab:accuracy_paradox} compares the two decomposers at the level of single steps.
The solver-blind decomposer reproduces the \ac{gt} subgoal two to three times as often as \ac{sad} in every setting.
Yet \ac{sad} reaches the higher synthesis accuracy in every setting.
In other words, the subgoals of \ac{sad} differ from the \ac{gt} subgoals more often but lead the synthesizer to the \ac{gt} subprogram more often.
This shows that a subgoal does not need to equal the \ac{gt} subgoal to lead the synthesizer to the \ac{gt} subprogram, and which subgoals do so depends on the inductive biases of the synthesizer.
\ac{sad} trades agreement with \ac{gt} for subgoals that this synthesizer resolves more reliably, so for this synthesizer, agreement with the \ac{gt} subgoal is a weak indicator of whether a subgoal is useful.
The decrease in decomposition accuracy also shows that the reward against the \ac{gt} program does not pull the decomposer back toward the \ac{gt} subgoals (Appendix~\ref{app:gtreward}).
\begin{table}[b]
\centering
\caption{Single-step decomposition and synthesis accuracy (\%) under teacher forcing, mean $\pm$ std.\ over 5 seeds. The solver-blind decomposer agrees more often with the \ac{gt} subgoals, whereas the subgoals of \ac{sad} more often lead the synthesizer to the \ac{gt} subprogram.}
\label{tab:accuracy_paradox}
\small
\begin{tabular}{llcccc}
\toprule
\textbf{Domain} & \textbf{Method} & \multicolumn{2}{c}{\textbf{Decomposition Acc.}} & \multicolumn{2}{c}{\textbf{Synthesis Acc.}} \\
 &  & Test & Length Gen. & Test & Length Gen. \\
\midrule
\multirow{2}{*}{Deepcoder}
 & Solver-blind & 79.2 $\pm$ 0.4 & 63.3 $\pm$ 0.6 & 73.0 $\pm$ 1.2 & 64.3 $\pm$ 0.8 \\
 & \ac{sad}              & 31.7 $\pm$ 0.7 & 21.5 $\pm$ 0.5 & 87.2 $\pm$ 1.3 & 79.8 $\pm$ 1.1 \\
\midrule
\multirow{2}{*}{Lambdabeam}
 & Solver-blind & 69.9 $\pm$ 0.7 & 48.5 $\pm$ 0.7 & 54.2 $\pm$ 0.5 & 41.3 $\pm$ 0.5 \\
 & \ac{sad}              & 33.7 $\pm$ 0.3 & 21.4 $\pm$ 0.3 & 61.9 $\pm$ 1.0 & 49.0 $\pm$ 1.2 \\
\bottomrule
\end{tabular}
\end{table}

\textbf{Generated decompositions} \quad
A beam oracle that selects the \ac{gt} decomposition wherever it appears in the beam improves task accuracy by less than 2 percentage points for both methods, and even with this oracle, the solver-blind baseline stays below \ac{sad} without it.
This shows that ranking the \ac{gt} subgoals higher does not close the gap, which therefore lies in the decompositions that each decomposer generates (Appendix~\ref{app:beamoracles}).

\begin{figure}[t]
     \centering
     \begin{subfigure}[b]{0.48\textwidth}
         \centering
         \includegraphics[width=\textwidth]{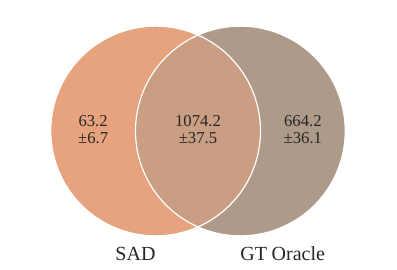}
         \caption{Deepcoder}
         \label{fig:venn_dc}
     \end{subfigure}
     \hfill
     \begin{subfigure}[b]{0.48\textwidth}
         \centering
         \includegraphics[width=\textwidth]{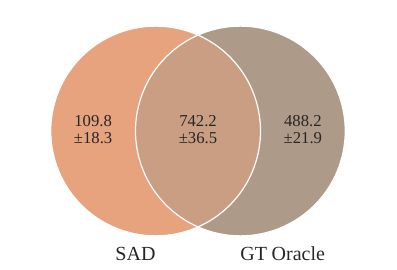}
         \caption{Lambdabeam}
         \label{fig:venn_lb}
     \end{subfigure}
     \caption{Tasks solved by \ac{sad} and by the \ac{gt} oracle, accumulated over the in-distribution and length-generalization test sets (mean $\pm$ std.\ over 5 seeds). The oracle solves more tasks overall, but a subset of tasks is solved only by \ac{sad}.}
     \label{fig:gtoracle}
\end{figure}

\textbf{\ac{sad} solves tasks on which \ac{gt} subgoals fail} \quad
\label{sec:gtoracle}
With \ac{gt} subgoals, the same synthesizer and search solve $94.7\% \pm 0.3\%$ and $79.8\% \pm 1.9\%$ of the tasks on Deepcoder (in-distribution and length generalization) and $89.6\% \pm 2.1\%$ and $51.3\% \pm 5.5\%$ on Lambdabeam.
These values are well above those of \ac{sad}, so a better decomposer still has considerable headroom.
At the task level, however, \ac{sad} solves $63.2 \pm 6.7$ tasks on Deepcoder and $109.8 \pm 18.3$ on Lambdabeam on which the oracle fails, counted over both settings with $2 \times 1000$ tasks per seed (Figure~\ref{fig:gtoracle}).
They differ from the \ac{gt} programs in length and primitive usage and fewer than 2 percentage points of these solutions match the \ac{gt} program on Deepcoder and none on Lambdabeam (Appendix~\ref{app:sadonly}).
This shows that for these tasks the \ac{gt} decomposition is not the best one for this synthesizer, and \ac{sad} reaches a different program instead, on Lambdabeam mostly a shorter one.
It does not show that the decompositions of \ac{sad} are better in general, as the oracle solves considerably more tasks.
The accuracy paradox thus extends from single subgoals to complete trajectories.

\textbf{No gain without decomposition ambiguity} \quad
\label{sec:robustfill}
On Robustfill, \ac{sad} and the solver-blind baseline reach the same task accuracy: $94.1\% \pm 0.3\%$ against $94.1\% \pm 0.4\%$ in-distribution and $90.8\% \pm 0.3\%$ against $90.8\% \pm 0.5\%$ under length generalization.
This is the outcome predicted in Section~\ref{sec:eval}: when the decomposition order is fixed, \ac{sad} performs like supervised training, and the additional \ac{rl} loss does not reduce accuracy.
It also indicates that the gains in the list domains do not stem from a generic effect of the additional loss, such as regularization, which would appear here as well, but from the choice among decompositions that Robustfill does not offer.
Overall, solver-aware training thus improves task accuracy where a task admits several decompositions and the synthesizer prefers some of them, and has no effect where the decomposition is fixed.

\section{Experiments with Pretrained \acp{llm}}
\label{sec:llm}
Section~\ref{sec:experiments} uses Transformers trained from scratch on the target \ac{dsl}; here we ask whether the effect holds when decomposer and synthesizer are a pretrained \ac{llm}.
The mismatch between \ac{gt} subgoals and the subgoals that the synthesizer can solve may be larger in this case, because the inductive biases of an \ac{llm} come mainly from pretraining and not from the \ac{dsl}.
Solver-aware training does not depend on where these biases come from, since it trains the decomposer against the synthesizer as it is.

\subsection{Evaluation}
\label{sec:llm_eval}
\textbf{Models and training} \quad
We use Llama-3.1-8B~\citep{grattafiori2024llama} with two LoRA adapters~\citep{hu2022lora}, one as decomposer and one as synthesizer.
Both adapters are fine-tuned on Deepcoder for 1000 steps, after which the synthesizer is frozen.
From the same decomposer checkpoint, we then train for 250 further steps in two ways: using \ac{sft} with $\mathcal{L}_{\text{sup}}$ only, which is the solver-blind control at equal budget, and with the full objective of Section~\ref{sec:sad_training} (\ac{sad}).
Both conditions use the same data and settings.

\textbf{Benchmark and baselines} \quad
We evaluate on the 200-task Deepcoder \ac{llm} benchmark of Exedec~\citep{shi2023exedec}, with at most 5 decomposition steps, one sample per prompt at temperature 0, and a single training seed.
Following Exedec, programs are written in a Pythonic version of the Deepcoder \ac{dsl}, in which each primitive except \texttt{Scanl1} becomes an equivalent Python one-liner (Appendix~\ref{app:llm_baselines}).
This representation is closer to the code that the \acp{llm} have seen during pretraining.
We evaluate in-distribution only, because we cannot control the pretraining distribution and thus cannot construct a length-generalization split that the model has not seen.
As reference, we evaluate the off-the-shelf Llama-3.1-8B with direct synthesis and with decomposition, where the same model predicts the subgoals and synthesizes the subprograms; Appendix~\ref{app:llm_baselines} adds two Qwen2.5 models.

\subsection{Results and Discussion}
\label{sec:llm_results}
\begin{table}[t]
\centering
\caption{Llama-3.1-8B on the Deepcoder \ac{llm} benchmark (in-distribution, single seed).}
\label{tab:llm_sad}
\small
\begin{tabular}{lcc}
\toprule
\textbf{Condition} & \textbf{Solved (of 200)} & \textbf{Solved (\%)} \\
\midrule
Off-the-shelf, direct synthesis & 5 & 2.5 \\
Off-the-shelf, with decomposition & 15 & 7.5 \\
\midrule
Fine-tuned starting checkpoint & 56 & 28.0 \\
\quad + 250 steps \ac{sft} (solver-blind) & 61 & 30.5 \\
\quad + 250 steps \ac{sad} & 69 & 34.5 \\
\bottomrule
\end{tabular}
\end{table}
Table~\ref{tab:llm_sad} reports the results.
Off-the-shelf, all three \acp{llm} solve at most $10\%$ of the tasks, with or without decomposition (Appendix~\ref{app:llm_baselines}), which shows that these models need fine-tuning on the \ac{dsl} to reach a useful accuracy.
After fine-tuning, the starting checkpoint solves $28.0\%$ of the tasks.
Further training improves this by $2.5$ points with \ac{sft} (9 tasks gained, 4 lost) and by $6.5$ points with \ac{sad} (17 gained, 4 lost), so that \ac{sad} is $4.0$ points ahead of \ac{sft} at equal budget (13 gained, 5 lost).
This is the same ordering as with the transformers trained from scratch, and it shows that further supervised training accounts for less than half of the gain of \ac{sad} over the starting checkpoint.
The results thus indicate that solver-aware training carries over to pretrained \acp{llm}, in line with the results of Section~\ref{sec:results}.

\section{Conclusion}
\label{sec:conclusion}
\ac{sad} trains the decomposer against a frozen learned synthesizer, with supervised training on \ac{gt} subgoals as a scaffold and \ac{scst} with the \ac{ce} of the synthesizer on the \ac{gt} program as reward.
Although the solver-blind decomposer agrees more often with the \ac{gt} subgoals, \ac{sad} reaches a higher synthesis and task accuracy on Deepcoder and Lambdabeam, under zero-shot transfer, and with a fine-tuned \ac{llm}, and it solves tasks on which the \ac{gt} subgoals fail; on Robustfill, where the decomposition order is fixed, both perform equally.
For this synthesizer, a subgoal that is faithful to \ac{gt} can thus be one that it cannot act on, and a subgoal that deviates from \ac{gt} can be one that it solves.
Decomposition quality therefore depends on the synthesizer that executes the subgoals, and the decomposer can be trained against it.

The same situation may arise wherever a learned model produces intermediate representations for a separately trained component with its own inductive biases.
Supervision with \ac{gt} intermediates then does not reflect what this component can execute, and training the producer against the component is one way to account for it.
Since \ac{sad} and the \ac{gt} oracle solve partly different tasks with different programs, decomposition diversity is a resource for future work, for example for ensembles of decomposers or for robustness under distribution shift.

\textbf{Limitations} \quad
\ac{sad} trains the decomposer for one frozen synthesizer, so it must be retrained when the synthesizer changes; joint training of both is not studied.
Each subgoal is rewarded for its own step only, so the granularity of the decomposition follows the \ac{gt} programs; multi-step rewards could let both models adapt the subtask size.
The reward needs a \ac{gt} program at training time, which is standard in \ac{pbe} but not always available, and a learned synthesizer: a search-based solver would only provide search success or cost, which we expect to be sparser, since it does not separate decompositions that all succeed, but have not tested.
Finally, the \ac{llm} experiment covers one model, one seed, and in-distribution tasks only.

\section*{Acknowledgments}
This work was supported by the German Federal Ministry for Research, Technology, and Space (BMFTR) and the German Federal Ministry for Economic Affairs and Energy (BMWE).
\bibliography{neurips_2026}
\bibliographystyle{plainnat}

\newpage
\appendix
\renewcommand{\thefigure}{\thesection.\arabic{figure}}
\renewcommand{\thetable}{\thesection.\arabic{table}}
\setcounter{figure}{0}
\setcounter{table}{0}
\section{Domains \& \acsp{dsl}}
\label{app:domains}

The three domains differ in program structure and state update, and therefore in how many valid decompositions a task admits.
Deepcoder and Lambdabeam admit several decompositions per task, whereas Robustfill fixes the decomposition order and serves as the negative control of the evaluation.

\subsection{Deepcoder}
Deepcoder tasks specify transformations of integer lists: the inputs of each \ac{io} example are one or two integer lists, possibly with scalar integers, and the output is an integer or a list.
A program is a sequence of lines, each of which applies a first- or higher-order operation such as \texttt{Map}, \texttt{Filter}, \texttt{ZipWith}, \texttt{Sort}, or \texttt{Scanl1}.
Its result becomes a new variable that later lines can use, and the last assigned variable is the output.
Figure~\ref{fig:dc_example_figure} shows an example task, and Figure~\ref{fig:dsl_dc} shows the \ac{dsl}.
\begin{figure}[h]
    \centering
    \begin{tikzpicture}[node distance=0.2cm,>=stealth,thick]
    \tikzstyle{io} = [rectangle, rounded corners, draw=black, fill=gray!20, inner sep=3pt, text width=6cm, align=center]
    
    \node[io] (pair1) {\textbf{I/O Pair 1}\\ Input: $\texttt{[5, 1, 6]}$, $\texttt{[-2, -15, 1]}$ \\ Output: $\texttt{[9, -30, 21]}$};
    
    \node[io, below=of pair1] (pair2) {\textbf{I/O Pair 2}\\ Input: $\texttt{[5]}$, $\texttt{[-4]}$ \\ Output: $\texttt{[3]}$};
    
    \node[io, below=of pair2] (pair3) {\textbf{I/O Pair 3}\\ Input: $\texttt{[10, 2]}$, $\texttt{[-2, -5]}$ \\ Output: $\texttt{[24, 15]}$};
    
    \node[align=left, below=of pair3, yshift=-0.3cm] (gt) {Ground Truth:\\$x_0 = \texttt{INPUT}$ \texttt{|} $x_1 = \texttt{INPUT}$ \texttt{|} $x_2 = \texttt{Scanl1 (max) } x_0$\\ \texttt{|} $x_3 = \texttt{ZipWith (+) x1 x2}$ \texttt{|} $x_4 = \texttt{Map (*3) } x_3$};
    
    \draw[dashed] 
      ($(pair1.north west)+(-0.1cm,0.1cm)$) 
      rectangle 
      ($(pair3.south east)+(0.1cm,-0.1cm)$);
    
    \end{tikzpicture}
    \caption{Example task from the Deepcoder domain. The program takes the running maximum of the first input, adds the second input, and multiplies the result by 3.}
    \label{fig:dc_example_figure}
    \end{figure}

\begin{figure}[h]
    \centering 
    \[
    \begin{array}{rl}
        \text{Program} \; P := & i_1; \; i_2; \; \ldots; \; a_1; \; a_2; \; \ldots \\[8pt]
        \text{Initialization} \; i := & v \gets \texttt{INPUT} \\[8pt]
        \text{Assignment} \; a := & v \gets f \;|\; v \gets h \\[8pt]
        \text{First-Order Operation} \; f := & \texttt{Head}(l) \;|\; \texttt{Last}(l) \;|\; \texttt{Access}(n, l) \;|\; \texttt{Minimum}(l) \;|\; \texttt{Maximum}(l) \\
                                              & \;|\; \texttt{Sum}(l) \;|\; \texttt{Take}(n, l) \;|\; \texttt{Drop}(n, l) \;|\; \texttt{Reverse}(l) \;|\; \texttt{Sort}(l) \\[8pt]
        \text{Higher-Order Operation} \; h := & \texttt{Map}(\lambda, l) \;|\; \texttt{Filter}(\beta, l) \;|\; \texttt{Count}(\beta, l) \;|\; \texttt{ZipWith}(\Sigma, l, l) \\
                                                 & \;|\; \texttt{Scanl1}(\Sigma, l) \\[8pt]
        \text{int} \to \text{int} \; \text{Lambda} \; \lambda := & (+1) \;|\; (-1) \;|\; (*2) \;|\; (/2) \;|\; (*(-1)) \;|\; (**2) \;|\; (*3) \;|\; (/3) \;|\; (*4) \;|\; (/4) \\[8pt]
        \text{int} \to \text{bool} \; \text{Lambda} \; \beta := & (> 0) \;|\; (< 0) \;|\; (\%2 == 0) \;|\; (\%2 == 1) \\[8pt]
        (\text{int}, \text{int}) \to \text{int} \; \text{Lambda} \; \Sigma := & (+) \;|\; (-) \;|\; (*) \;|\; (\texttt{min}) \;|\; (\texttt{max}) \\[8pt]
        \text{Integer Variable} \; n := & v \\[8pt]
        \text{List Variable} \; l := & v \\[8pt]
        \text{Variable Name} \; v := & x_0 \;|\; x_1 \;|\; \ldots \\
    \end{array}
    \]
    \caption{\ac{dsl} of the Deepcoder domain.}
    \label{fig:dsl_dc}
\end{figure}

Since later lines can use all earlier results, the task does not fix the structure of the program, and several programs and decompositions usually produce the same outputs.
Program length is the number of non-input lines, i.e., the number of decomposition steps.

For in-distribution evaluation, we train and test on programs of length 1--4; for length generalization, we train on lengths 1--4 and test on length 5.
Each task has 3 \ac{io} examples plus one hidden example, and the input lists have at most 5 elements.
A task generated by sampling a program of length 5 can have an equivalent shorter program of a training length, so it would not test length generalization.
We therefore enumerate all programs up to the maximum length and record all minimal-length solutions of each task.
A training task needs at least one minimal-length solution of a training length, and a length-generalization test task needs all its minimal-length solutions to be of the test length, so no test task can be solved by a program of a training length.

\subsection{Lambdabeam}
Lambdabeam extends Deepcoder in two ways.
First, lambda functions are built during synthesis: instead of selecting from a fixed set of lambdas such as \texttt{(+1)} or \texttt{(*2)}, the synthesizer chooses the operation and its argument, which is either an integer constant in $[-5, 5]$ or a program variable that holds such a constant.
Second, a conditional \texttt{If} selects element-wise between two lists, depending on a condition on a third list.
Apart from the \texttt{Access} operation, which Lambdabeam does not contain, the remaining operations, the state representation, and the state update are the same as in Deepcoder.
Figure~\ref{fig:lb_example_figure} shows an example task, and Figure~\ref{fig:dsl_lb} shows the rules of the \ac{dsl} that differ from Deepcoder.
\begin{figure}[h]
\centering
\begin{tikzpicture}[node distance=0.2cm,>=stealth,thick]
\tikzstyle{io} = [rectangle, rounded corners, draw=black, fill=gray!20, inner sep=3pt, text width=6cm, align=center]

\node[io] (pair1) {\textbf{I/O Pair 1}\\ Input: $\texttt{[5, 6]}$ \\ Output: $\texttt{[7, 0]}$};

\node[io, below=of pair1] (pair2) {\textbf{I/O Pair 2}\\ Input: $\texttt{[4, 1]}$ \\ Output: $\texttt{[6, 3]}$};

\node[io, below=of pair2] (pair3) {\textbf{I/O Pair 3}\\ Input: $\texttt{[5, 4, 2]}$ \\ Output: $\texttt{[7, 1, 0]}$};

\node[align=left, below=of pair3, yshift=-0.3cm] (gt) {Ground Truth:\\$x_0 = \texttt{INPUT}$ \texttt{|} $x_1 = \texttt{Map Subtract -2 x0}$ \texttt{|} $x_2 = \texttt{Scanl1 IntDivide } x_1$\\ \texttt{|} $x_3 = \texttt{If Less 2}$ $x_0$ $x_1$ $x_2$};

\draw[dashed] 
  ($(pair1.north west)+(-0.1cm,0.1cm)$) 
  rectangle 
  ($(pair3.south east)+(0.1cm,-0.1cm)$);

\end{tikzpicture}
\caption{Example task from the Lambdabeam domain. The program adds 2 to each element, computes a running integer division, and keeps the shifted value wherever the input element is less than 2.}
\label{fig:lb_example_figure}
\end{figure}
\begin{figure}[h]
    \centering
    \[
        \begin{array}{rl}
            \text{Assignment}\; a := & v \gets f\;|\; v \gets h\;|\; v \gets c \\[8pt]
            \text{Conditional Operation}\; c := & \texttt{If}(\beta, l, l, l) \\[8pt]
            \text{int} \to \text{int}\; \text{Lambda}\; \lambda := & (+\; e)\;|\; (-\; e)\;|\; (*\; e)\;|\; (/\; e) \\[8pt]
            \text{int} \to \text{bool}\; \text{Lambda}\; \beta := & (>\; e)\;|\; (<\; e)\;|\; (\%2 == 0)\;|\; (\%2 == 1) \\[8pt]
            \text{Lambda Argument}\; e := & -5 \;|\; \ldots \;|\; 5 \;|\; n \\
        \end{array}
    \]
\caption{Rules of the Lambdabeam \ac{dsl} that differ from the Deepcoder \ac{dsl} (Figure~\ref{fig:dsl_dc}) or are missing there. All other rules are the same, except that Lambdabeam has no \texttt{Access} operation.}
\label{fig:dsl_lb}
\end{figure}
Both extensions enlarge the set of candidate subprograms per step.
A subgoal that leads the synthesizer to an unusual lambda or an unnecessary \texttt{If} is therefore more likely to fall outside the beam than in Deepcoder.
Splits and data generation are the same as for Deepcoder.

\subsection{Robustfill}
Robustfill tasks map one input string to one output string with a \ac{dsl} of substring, case, token-extraction, and composition operations.
A program is a concatenation of expressions, each of which is applied to the original input and appends its result to the output.
Figure~\ref{fig:string_example_figure} shows an example task, and Figure~\ref{fig:dsl_rf} shows the \ac{dsl}.
\begin{figure}[b]
\centering
\begin{tikzpicture}[node distance=0.2cm,>=stealth,thick]
\tikzstyle{io} = [rectangle, rounded corners, draw=black, fill=gray!20, inner sep=3pt, text width=9cm, align=center]

\node[io] (pair1) {\textbf{I/O Pair 1}\\ Input: \texttt{alan Turing1} \\ Output: \texttt{1.TURING,Alan}};
\node[io, below=of pair1] (pair2) {\textbf{I/O Pair 2}\\ Input: \texttt{21.Donald@knuTh} \\ Output: \texttt{21.KNUTH,Donald}};
\node[io, below=of pair2] (pair3) {\textbf{I/O Pair 3}\\ Input: \texttt{8:grace,HoppER} \\ Output: \texttt{8.HOPPER,Grace}};
\node[io, below=of pair3] (pair4) {\textbf{I/O Pair 4}\\ Input: \texttt{EDSGER99 DIJKSTRA} \\ Output: \texttt{99.DIJKSTRA,Edsger}};

\node[align=left, below=of pair4, yshift=-0.3cm] (gt) {Ground Truth: 
$\texttt{GetAll(NUMBER) | ConstStr('.') |}$ \\ $\texttt{Compose(ToCase(ALL\_CAPS), GetToken(WORD, -1)) |}$\\
$\texttt{ConstStr(',') | Compose(ToCase(PROPER), GetToken(WORD, 1))}$};

\draw[dashed] 
  ($(pair1.north west)+(-0.1cm,0.1cm)$) 
  rectangle 
  ($(pair4.south east)+(0.1cm,-0.1cm)$);

\end{tikzpicture}
\caption{Example task from the Robustfill domain. The program outputs the number, then the last name in capitals, and then the first name in title case.}
\label{fig:string_example_figure}
\end{figure}
\begin{figure}[!b]
    \centering
    \[
    \begin{array}{rl}
        \text{Program} \; P := & \texttt{Concat}(e_1, e_2, \ldots) \\
        \text{Expression} \; e := & s \;|\; m \;|\; o \;|\; \texttt{ConstStr}(c) \\
        \text{Compose} \; o := & m_1(m_2) \;|\; m(s) \\
        \text{Substring} \; s := & \texttt{SubStr}(k_1, k_2) \;|\; \texttt{GetSpan}(r_1, i_1, b_1, r_2, i_2, b_2) \\
                                    & \;|\; \texttt{GetUpto}(r, i) \;|\; \texttt{GetFrom}(r, i) \;|\; \texttt{GetToken}(r, i) \\
        \text{Modification} \; m := & \texttt{ToCase}(a) \;|\; \texttt{Replace}(c_1, c_2) \;|\; \texttt{Trim()} \\
                                      & \;|\; \texttt{GetFirst}(r, i) \;|\; \texttt{GetAll}(r) \\
                                      & \;|\; \texttt{Substitute}(r, i, c) \;|\; \texttt{SubstituteAll}(r, c) \\
                                      & \;|\; \texttt{Remove}(r, i) \;|\; \texttt{RemoveAll}(r) \\
        \text{Regex} \; r := & \texttt{NUMBER} \;|\; \texttt{WORD} \;|\; \texttt{ALPHANUM} \;|\; \texttt{ALL\_CAPS} \;|\; \texttt{PROPER\_CASE} \\
                                & \;|\; \texttt{LOWER} \;|\; \texttt{DIGIT} \;|\; \texttt{CHAR} \;|\; \delta \\
        \text{Case} \; a := & \texttt{ALL\_CAPS} \;|\; \texttt{PROPER\_CASE} \;|\; \texttt{LOWER} \\
        \text{Position} \; k := & -100 \;|\; -99 \;|\; \ldots \;|\; -1 \;|\; 0 \;|\; 1 \;|\; 2 \;|\; \ldots \;|\; 100 \\
        \text{Index} \; i := & -5 \;|\; -4 \;|\; \ldots \;|\; -1 \;|\; 1 \;|\; 2 \;|\; \ldots \;|\; 5 \\
        \text{Boundary} \; b := & \texttt{START} \;|\; \texttt{END} \\
        \text{Character} \; c := & A \;|\; \ldots \;|\; Z \;|\; a \;|\; \ldots \;|\; z \;|\; 0 \;|\; \ldots \;|\; 9 \;|\; \delta \\
        \text{Delimiter} \; \delta := & \texttt{\& , . ?  ! @ ()  [] \% \# \$ " ´} \\
    \end{array}
    \]
    \caption{\ac{dsl} of the Robustfill domain.}
    \label{fig:dsl_rf}
\end{figure}

The state update differs from the list domains.
After each step, its output is removed from the front of the remaining target, so the residual shows what is left to produce, and the subtasks become more specified as the trajectory advances.
Since each step must produce a prefix of the residual, the decomposition order is fixed: a step whose output is not such a prefix cannot be continued.
The implementation of a single step can still vary, but the sequence of subtasks is determined by the task.

Without alternative decompositions, solver feedback has nothing to choose between, so \ac{sad} should perform like supervised training in this domain.
Robustfill is therefore the negative control of the evaluation.

For in-distribution evaluation, we train and test on programs of length 1--6; for length generalization, we train on lengths 1--6 and test on lengths 7--10.
Each task has 4 \ac{io} examples plus one hidden example, and the input strings have at most 20 characters.
A program is sampled from the target length range such that it executes on all inputs, and its outputs form the examples.
Each concatenated expression is a subprogram, and program length is the number of subprograms.

\setcounter{figure}{0}
\setcounter{table}{0}
\section{Failure cases}
\label{app:failure_cases}
This appendix shows two Deepcoder tasks on which \ac{sad} succeeds and a method that uses \ac{gt} subgoals fails.
In the first, the solver-blind decomposer proposes a subgoal that no single \ac{dsl} operation produces; in the second, every \ac{gt} subgoal is realizable, but the \ac{gt} oracle still does not solve the task.

\subsection{Subgoal Outside the \ac{dsl}}
The task in Figure~\ref{fig:struct_mismatch_example} maps two input lists to one output list with a \ac{gt} program of four steps, and both decomposers first predict the subgoal for \texttt{x2}.
No single Deepcoder operation produces the subgoal of the solver-blind decomposer.
In the second example, the value \texttt{1} must become \texttt{1} at the first position and \texttt{16} at the last, which rules out \texttt{Map} and \texttt{ZipWith} of \texttt{x0} with itself.
\texttt{ZipWith} with \texttt{x1} shortens the list to the length of \texttt{x1} and \texttt{Scanl1} keeps the first element, which the third example changes from \texttt{3} to \texttt{16}.
The remaining list operations only select or reorder existing values, none of which is \texttt{16}, and all other operations return a single integer.
\begin{figure}[!b]
  \centering
  \begin{tikzpicture}[
    font=\small,
    every node/.style={inner sep=0pt},
  ]
  
  \node[
    draw=specColor,
    fill=specfill,
    rounded corners=6pt,
    inner sep=8pt,
    text width=11.8cm,
    align=center
  ] (taskspec) {
    \textbf{\textcolor{specColor}{Task specification}}\\[4pt]
    \begin{tabular}{p{5.4cm} p{0.1cm} p{5.0cm}}
      \centering\textcolor{specColor}{Input} & &
      \centering\textcolor{specColor}{Output}\tabularnewline
      \midrule
      \centering\textcolor{specColor}{\texttt{x0=[], x1=[0]}} & &
      \centering\textcolor{specColor}{\texttt{[]}}\tabularnewline[2pt]
      \centering\textcolor{specColor}{\texttt{x0=[1,0,6,9,1], x1=[9]}} & &
      \centering\textcolor{specColor}{\texttt{[1, 0, 22, 45, 1]}}\tabularnewline[2pt]
      \centering\textcolor{specColor}{\texttt{x0=[3,7,1,4], x1=[-3,-1]}} & &
      \centering\textcolor{specColor}{\texttt{[7, 25, 1, 9]}}\tabularnewline
    \end{tabular}\\[5pt]
    \textcolor{specColor}{\rule{\linewidth}{0.4pt}}\\[3pt]
    \textcolor{specColor}{\texttt{GT:}}~
    \textcolor{specColor}{\texttt{x0=INPUT | x1=INPUT | x2=Map(/3) x0 | x3=ZipWith(+) x0 x2}}\\
    \textcolor{specColor}{\texttt{x4=ZipWith(*) x2 x3 | x5=ZipWith(+) x0 x4}}
  };
  
  \node[
    draw=blindColor,
    fill=white,
    fill opacity=0.6,
    rounded corners=6pt,
    inner sep=6pt,
    text width=5.6cm,
    below left=0.5cm and 0.0cm of taskspec.south,
    anchor=north east,
    xshift=-0.1cm
  ] (agnosticbox) {
    \textbf{\textcolor{blindColor!15!black}{Solver-blind}}\\
    \textcolor{specColor!15!black}{\small predicted subgoal (step 1)}\\[6pt]
    \textcolor{specColor}{\small Predicted \texttt{x2}}\\[4pt]
    \colorbox{subtaskbad}{\parbox{\dimexpr\linewidth-2\fboxsep}{%
      \centering\textcolor{blindColor!15!black}{\texttt{[] $\to$ []}}
    }}\\[3pt]
    \colorbox{subtaskbad}{\parbox{\dimexpr\linewidth-2\fboxsep}{%
      \centering\textcolor{blindColor!15!black}{\texttt{[1,0,6,9,1] $\to$ [1,0,16,16,16]}}
    }}\\[3pt]
    \colorbox{subtaskbad}{\parbox{\dimexpr\linewidth-2\fboxsep}{%
      \centering\textcolor{blindColor!15!black}{\texttt{[3,7,1,4] $\to$ [16,16,16,16]}}
    }}\\[6pt]
    \tikz\node[
      draw=blindColor, fill=blindColor!20!white,
      rounded corners=4pt, inner sep=4pt,
      text width=\linewidth-8pt, align=center
    ]{\textbf{\textcolor{blindColor!15!black}{No single operation produces it~$\times$}}};
  };
  
  \node[
    draw=sadColor,
    fill=white,
    fill opacity=0.6,
    rounded corners=6pt,
    inner sep=6pt,
    text width=5.6cm,
    below right=0.5cm and 0.0cm of taskspec.south,
    anchor=north west,
    xshift=0.1cm
  ] (sadbox) {
    \textbf{\textcolor{sadColor!15!black}{SAD}}\\
    \textcolor{specColor!15!black}{\small predicted subgoal (step 1)}\\[6pt]
    \textcolor{specColor}{\small Predicted \texttt{x2}}\\[4pt]
    \colorbox{subtaskgood}{\parbox{\dimexpr\linewidth-2\fboxsep}{%
      \centering\textcolor{sadColor!15!black}{\texttt{[] $\to$ []}}
    }}\\[3pt]
    \colorbox{subtaskgood}{\parbox{\dimexpr\linewidth-2\fboxsep}{%
      \centering\textcolor{sadColor!15!black}{\texttt{[1,0,6,9,1] $\to$ [0,0,2,3,0]}}
    }}\\[3pt]
    \colorbox{subtaskgood}{\parbox{\dimexpr\linewidth-2\fboxsep}{%
      \centering\textcolor{sadColor!15!black}{\texttt{[3,7,1,4] $\to$ [1,2,0,1]}}
    }}\\[6pt]
    \tikz\node[
      draw=sadColor, fill=sadColor!20!white,
      rounded corners=4pt, inner sep=4pt,
      text width=\linewidth-8pt, align=center
    ]{\textbf{\textcolor{sadColor!15!black}{GT subgoal, solved by \texttt{Map(/3) x0}~$\checkmark$}}};
  };
  
  \end{tikzpicture}
  \caption{First decomposition step on a Deepcoder task. No single \ac{dsl} operation produces the subgoal of the solver-blind decomposer, so synthesis fails at this step. \ac{sad} predicts the \ac{gt} subgoal, which the synthesizer solves with \texttt{Map(/3) x0}, and \ac{sad} solves the task.}
  \label{fig:struct_mismatch_example}
\end{figure}
The synthesizer thus receives a subtask without a consistent solution, and the solver-blind method does not solve the task.
\ac{sad} predicts the \ac{gt} subgoal, which the synthesizer solves with \texttt{Map(/3) x0}, and the remaining steps follow the \ac{gt} program to a correct solution.
This shows a kind of error that the reward of \ac{sad} penalizes directly: since no subprogram is consistent with such a subgoal, the synthesizer is unlikely to assign a high probability to the \ac{gt} subprogram, whereas $\mathcal{L}_{\text{sup}}$ only measures the distance of the subgoal to the \ac{gt} subgoal.

\subsection{Realizable \ac{gt} Path That the Synthesizer Does Not Follow}
The task in Figure~\ref{fig:proc_mismatch_example} shows a failure that remains when the decomposition is perfect.
The \ac{gt} oracle receives every \ac{gt} subgoal at inference, and all of them are realizable in the \ac{dsl}, yet neither the oracle nor the solver-blind method solves the task.
The failure therefore lies in synthesis along the \ac{gt} path, which takes five steps through \texttt{Scanl1(-)} and three \texttt{ZipWith(-)} operations.
\ac{sad} solves the task in four steps with a different program, which multiplies \texttt{x0} element-wise with the halved \texttt{x1}, negates the result, and takes the element-wise maximum with \texttt{x1}.
This path avoids \texttt{Scanl1} and the chain of subtractions, which is possibly where the synthesizer fails on the \ac{gt} path.
The example matches the aggregate analysis in Appendix~\ref{app:sadonly}, where fewer than $2\%$ of the \ac{sad}-only solutions on Deepcoder match the \ac{gt} program and their primitive usage differs from \ac{gt}; a single task, however, does not show which primitives are hard for the synthesizer in general.
Training decomposers to find such alternative decompositions deliberately is left for future work.
\begin{figure}[h]
    \centering
    \begin{tikzpicture}[
      font=\small,
      every node/.style={inner sep=0pt},
    ]

    \node[
      draw=specColor,
      fill=specfill,
      rounded corners=6pt,
      inner sep=8pt,
      text width=11.8cm,
      align=center
    ] (taskspec) {
      \textbf{\textcolor{specColor}{Task specification}}\\[4pt]
      \begin{tabular}{p{5.4cm} p{0.1cm} p{5.0cm}}
        \centering\textcolor{specColor}{Input} & &
        \centering\textcolor{specColor}{Output}\tabularnewline
        \midrule
        \centering\textcolor{specColor}{\texttt{x0=[], x1=[0]}} & &
        \centering\textcolor{specColor}{\texttt{[]}}\tabularnewline[2pt]
        \centering\textcolor{specColor}{\texttt{x0=[1,0,6,9,1], x1=[9]}} & &
        \centering\textcolor{specColor}{\texttt{[9]}}\tabularnewline[2pt]
        \centering\textcolor{specColor}{\texttt{x0=[3,7,1,4], x1=[-3,-1]}} & &
        \centering\textcolor{specColor}{\texttt{[6, 7]}}\tabularnewline
      \end{tabular}
    };

    \node[
      draw=gtOracleColor,
      fill=white,
      fill opacity=0.6,
      rounded corners=6pt,
      inner sep=6pt,
      text width=5.6cm,
      below left=0.5cm and 0.0cm of taskspec.south,
      anchor=north east,
      xshift=-0.1cm
    ] (gtbox) {
      \textbf{\textcolor{gtOracleColor!15!black}{GT solution}}\\
      \textcolor{specColor!15!black}{\small (5 steps)}\\[6pt]
      \colorbox{specfill}{\parbox{\dimexpr\linewidth-2\fboxsep}{%
        \centering\textcolor{gtOracleColor!15!black}{\texttt{x0 = INPUT}}
      }}\\[3pt]
      \colorbox{specfill}{\parbox{\dimexpr\linewidth-2\fboxsep}{%
        \centering\textcolor{gtOracleColor!15!black}{\texttt{x1 = INPUT}}
      }}\\[3pt]
      \colorbox{gtOracleColor!12!white}{\parbox{\dimexpr\linewidth-2\fboxsep}{%
        \centering\textcolor{gtOracleColor!15!black}{\texttt{x2 = ZipWith(-) x0 x1}}
      }}\\[3pt]
      \colorbox{gtOracleColor!12!white}{\parbox{\dimexpr\linewidth-2\fboxsep}{%
        \centering\textcolor{gtOracleColor!15!black}{\texttt{x3 = Scanl1(-) x1}}
      }}\\[3pt]
      \colorbox{gtOracleColor!12!white}{\parbox{\dimexpr\linewidth-2\fboxsep}{%
        \centering\textcolor{gtOracleColor!15!black}{\texttt{x4 = ZipWith(-) x1 x3}}
      }}\\[3pt]
      \colorbox{gtOracleColor!12!white}{\parbox{\dimexpr\linewidth-2\fboxsep}{%
        \centering\textcolor{gtOracleColor!15!black}{\texttt{x5 = ZipWith(max) x1 x2}}
      }}\\[3pt]
      \colorbox{gtOracleColor!12!white}{\parbox{\dimexpr\linewidth-2\fboxsep}{%
        \centering\textcolor{gtOracleColor!15!black}{\texttt{x6 = ZipWith(-) x5 x4}}
      }}\\[6pt]
      \tikz\node[
        draw=gtOracleColor, fill=gtOracleColor!20!white,
        rounded corners=4pt, inner sep=4pt,
        text width=\linewidth-8pt, align=center
      ]{\textbf{\textcolor{gtOracleColor!15!black}{Output: x6~$\checkmark$}}};
    };

    \node[
      draw=sadColor,
      fill=white,
      fill opacity=0.6,
      rounded corners=6pt,
      inner sep=6pt,
      text width=5.6cm,
      below right=0.5cm and 0.0cm of taskspec.south,
      anchor=north west,
      xshift=0.1cm
    ] (sadbox) {
      \textbf{\textcolor{sadColor!15!black}{SAD solution}}\\
      \textcolor{specColor!15!black}{\small (4 steps)}\\[6pt]
      \colorbox{specfill}{\parbox{\dimexpr\linewidth-2\fboxsep}{%
        \centering\textcolor{sadColor!15!black}{\texttt{x0 = INPUT}}
      }}\\[3pt]
      \colorbox{specfill}{\parbox{\dimexpr\linewidth-2\fboxsep}{%
        \centering\textcolor{sadColor!15!black}{\texttt{x1 = INPUT}}
      }}\\[3pt]
      \colorbox{sadColor!12!white}{\parbox{\dimexpr\linewidth-2\fboxsep}{%
        \centering\textcolor{sadColor!15!black}{\texttt{x2 = Map(/2) x1}}
      }}\\[3pt]
      \colorbox{sadColor!12!white}{\parbox{\dimexpr\linewidth-2\fboxsep}{%
        \centering\textcolor{sadColor!15!black}{\texttt{x3 = ZipWith(*) x0 x2}}
      }}\\[3pt]
      \colorbox{sadColor!12!white}{\parbox{\dimexpr\linewidth-2\fboxsep}{%
        \centering\textcolor{sadColor!15!black}{\texttt{x4 = Map(*(-1)) x3}}
      }}\\[3pt]
      \colorbox{sadColor!12!white}{\parbox{\dimexpr\linewidth-2\fboxsep}{%
        \centering\textcolor{sadColor!15!black}{\texttt{x5 = ZipWith(max) x1 x4}}
      }}\\[3pt]
      
      \phantom{\colorbox{sadColor!12!white}{\parbox{\dimexpr\linewidth-2\fboxsep}{%
        \centering\textcolor{sadColor!15!black}{\texttt{x6 = ZipWith(-) x5 x4}}
      }}}\\[6pt]
      \tikz\node[
        draw=sadColor, fill=sadColor!20!white,
        rounded corners=4pt, inner sep=4pt,
        text width=\linewidth-8pt, align=center
      ]{\textbf{\textcolor{sadColor!15!black}{Output: x5~$\checkmark$}}};
    };

    \end{tikzpicture}
    \caption{\ac{gt} program (5 steps) and \ac{sad} solution (4 steps) for a Deepcoder task; both satisfy the \ac{io} examples. With the \ac{gt} subgoals, the \ac{gt} oracle does not solve the task, and neither does the solver-blind method. \ac{sad} solves it with a different program that avoids \texttt{Scanl1} and the chain of \texttt{ZipWith(-)} operations.}
    \label{fig:proc_mismatch_example}
\end{figure}

\setcounter{figure}{0}
\setcounter{table}{0}
\section{Training and Inference Setup}
\label{app:train_and_inf}
\ac{sad} and the solver-blind baseline share architecture, hyperparameters, training data, synthesizer weights, and search, and differ only in the training signal of the decomposer.

\subsection{Training Setup}
\label{app:training}
Synthesizer and decomposer are 3-layer transformers with embedding dimension 512 and hidden dimension 1024.
The synthesizer is trained with learning rate $2\times10^{-4}$ and the decomposer with $1\times10^{-4}$, both with square-root decay after 16k linear warmup steps and for 500k steps in total.
The batch size is 128 for the list domains and 64 for Robustfill, to fit into GPU memory.

\paragraph{Synthesizer training.}
All synthesizers are trained with teacher forcing to generate the next \ac{gt} subprogram $p^*_t$ and are frozen afterwards, so that the synthesizer stays fixed during decomposer training.
They differ in the input they receive during training.
The synthesizer of Exedec receives the \ac{gt} subtask $(I^{(t)}, O^{*}_t)$ and thus learns to implement a decomposition step that is already given.
The synthesizer of \ac{sad} and the solver-blind baseline receives the full current task $s_t = (I^{(t)}, O^{(t)})$, without a subgoal, and thus has to infer which subtask comes next as well as how to implement it.
This makes it informative also on subgoals that leave the \ac{gt} trajectory, which are the subgoals that the \ac{scst} reward has to score (Section~\ref{sec:synth_training}).
At inference, all three methods give the synthesizer the subtask $(I^{(t)}, O^{(t)}_{\text{pred}})$ induced by the predicted subgoal (Algorithm~\ref{algo:synth}), and the \ac{scst} reward uses the same input.

\paragraph{Decomposer training and compared methods.}
Exedec and the solver-blind baseline train the decomposer with $\mathcal{L}_{\text{sup}}$ only, with identical settings.
\ac{sad} trains it with the full objective of Section~\ref{sec:sad_training} from the start of training, without a supervised warm-up phase or a pretrained checkpoint.
The three methods thus combine the two components as follows:
\begin{itemize}
    \item \textbf{Exedec}: synthesizer trained on \ac{gt} subtasks, decomposer trained with $\mathcal{L}_{\text{sup}}$.
    \item \textbf{Solver-blind}: synthesizer trained on the full current task, decomposer trained with $\mathcal{L}_{\text{sup}}$.
    \item \textbf{\ac{sad}}: synthesizer trained on the full current task, decomposer trained with $\mathcal{L}_{\text{sup}}$ and \ac{scst}.
\end{itemize}
Each adjacent pair differs in one component.
\ac{sad} and the solver-blind baseline differ only in the training signal of the decomposer, so their comparison shows the effect of solver-aware training, and the conclusions in Section~\ref{sec:results} rest on it.
Exedec and the solver-blind baseline differ only in the synthesizer, so their comparison shows the effect of training the synthesizer on the full current task.
Both \ac{sad} and the solver-blind baseline are trained on single decomposition steps; the larger gains of \ac{sad} on longer programs (Appendix~\ref{app:perfvslen}) thus arise without a multi-step objective.

\paragraph{Compute.}
Training the synthesizer and training the decomposer with $\mathcal{L}_{\text{sup}}$ each take 5--6 hours per seed on a single NVIDIA RTX PRO 6000.
\ac{sad} training takes about $30\%$ longer than supervised decomposer training, because the reward needs two additional forward passes of the frozen synthesizer per training state, one for the sampled and one for the greedy subgoal.
Inference cost is the same for all methods.

\subsection{Beam Search}
\label{app:beamsearch}
Inference follows the beam search of Exedec~\cite{shi2023exedec}.
A single beam of $k = 10$ candidate trajectories runs through the whole synthesis process and is extended in place at every model call, without restarting at each decomposition or synthesis step.
Each candidate is a partial solution, that is, a sequence of predicted subgoals and the subprograms synthesized for them.
Candidates are ranked by the sum of the log-probabilities of the decomposer and the synthesizer, and they are pruned if they fail to parse or execute, are functionally redundant with a higher-ranked candidate, or exceed domain-specific computational limits.
The search stops after at most $T = 10$ steps for the list domains and $T = 20$ for Robustfill.
The \ac{gt} oracle uses the same beam search with the \ac{gt} subgoals in place of the predicted ones.

\setcounter{figure}{0}
\setcounter{table}{0}
\section{Additional Experiments and Analyses with Transformers}
\subsection{Statistical Significance Tests}
\label{app:pvalues}
We compare the task accuracies of the three methods with two-sided paired $t$-tests over the 5 seeds, which gives 4 degrees of freedom, at a significance level of 5\%.
Table~\ref{tab:pvalues} reports the results.
\ac{sad} is significantly better than the solver-blind baseline in all four settings of the list domains, and the two methods do not differ significantly on Robustfill.
Against Exedec, \ac{sad} is significantly better in the list domains except on Deepcoder in-distribution, while on Robustfill in-distribution, Exedec is significantly better than both other methods. 
However, the effect is minimal there.
\begin{table}[h]
\centering
\caption{Paired $t$-tests on task accuracy over 5 seeds, reported as $p$-value ($t$-statistic). Bold $p$-values are significant at the 5\% level.}
\label{tab:pvalues}
\small
\setlength{\tabcolsep}{4pt}
\begin{tabular}{llccc}
\toprule
\textbf{Domain} & \textbf{Setting} & \textbf{\ac{sad} vs.\ solver-blind} & \textbf{Solver-blind vs.\ Exedec} & \textbf{\ac{sad} vs.\ Exedec} \\
\midrule
\multirow{2}{*}{Deepcoder}
& In-distribution & $\mathbf{7.68\times10^{-3}}$ ($5.00$) & $5.18\times10^{-2}$ ($-2.74$) & $1.77\times10^{-1}$ ($1.63$) \\
& Length gen. & $\mathbf{6.88\times10^{-3}}$ ($5.12$) & $\mathbf{3.24\times10^{-2}}$ ($3.22$) & $\mathbf{2.18\times10^{-5}}$ ($22.84$) \\
\addlinespace
\multirow{2}{*}{Lambdabeam}
& In-distribution & $\mathbf{2.60\times10^{-2}}$ ($3.45$) & $\mathbf{2.23\times10^{-2}}$ ($3.62$) & $\mathbf{1.34\times10^{-4}}$ ($7.98$) \\
& Length gen. & $\mathbf{7.65\times10^{-3}}$ ($4.97$) & $\mathbf{5.76\times10^{-4}}$ ($9.94$) & $\mathbf{5.35\times10^{-4}}$ ($10.13$) \\
\addlinespace
\multirow{2}{*}{Robustfill}
& In-distribution & $8.76\times10^{-1}$ ($0.17$) & $\mathbf{3.57\times10^{-2}}$ ($-3.12$) & $\mathbf{1.31\times10^{-2}}$ ($-4.26$) \\
& Length gen. & $6.14\times10^{-1}$ ($-0.55$) & $2.00\times10^{-1}$ ($-1.53$) & $9.53\times10^{-2}$ ($-2.18$) \\
\bottomrule
\end{tabular}
\end{table}

\subsection{Performance Gap Grows with Program Length}
\label{app:perfvslen}
\begin{figure}[b]
     \centering
     \begin{subfigure}[b]{0.48\textwidth}
         \centering
         \includegraphics[width=\textwidth]{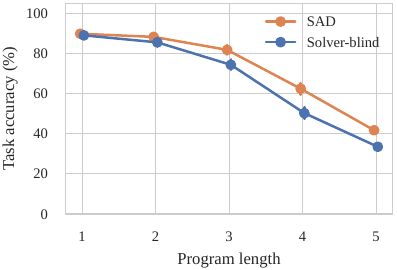}
         \caption{Deepcoder}
         \label{fig:perflen_dc}
     \end{subfigure}
     \hfill
     \begin{subfigure}[b]{0.48\textwidth}
         \centering
         \includegraphics[width=\textwidth]{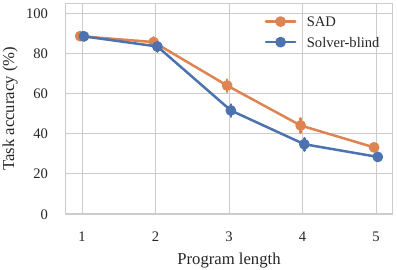}
         \caption{Lambdabeam}
         \label{fig:perflen_lb}
     \end{subfigure}
     \caption{Task accuracy over the length of the \ac{gt} program in the list domains. The advantage of \ac{sad} over the solver-blind baseline grows with program length, although both decomposers are trained on single steps.}
     \label{fig:perflen_lists}
\end{figure}

In both list domains, the advantage of \ac{sad} over the solver-blind baseline grows with the length of the \ac{gt} program (Figure~\ref{fig:perflen_lists}), although both decomposers are trained on single decomposition steps.
This indicates that the benefit of solver-aware subgoals accumulates over the steps of a task.
The state update of the list domains is a likely reason: a subgoal that the synthesizer cannot act on uses up one of the $T$ steps and adds its result to the state on which all later steps are conditioned, so each such subgoal also affects the remaining steps, and longer programs offer more occasions for it.
\begin{figure}[!b]
    \centering
    \includegraphics[width=.48\textwidth]{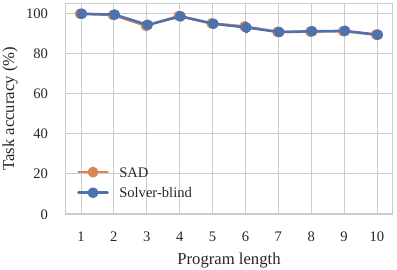}
     \caption{Task accuracy over the length of the \ac{gt} program on Robustfill. Accuracy decreases slowly for all methods, and \ac{sad} and the solver-blind baseline perform equally at every length.}
     \label{fig:perflen_rf}
\end{figure}

On Robustfill (Figure~\ref{fig:perflen_rf}), accuracy decreases more slowly with program length for all methods, and \ac{sad} and the solver-blind baseline perform equally at every length, as predicted for a domain without decomposition ambiguity.
Here, each step removes its output from the front of the remaining target string, so the residual shows what is left to produce, and the task likely becomes easier with each step.
In the list domains, by contrast, the target stays fixed while intermediate results accumulate as inputs, so the remaining target does not show which operations are still needed.
This possibly also explains why accuracy decreases more steeply with program length in the list domains for all methods.

\subsection{\texorpdfstring{$\mathcal{L}_{\text{sup}}$}{L\_sup} Ablation}
\label{app:lsupablation}
\begin{figure}[h]
     \centering
     \begin{subfigure}{0.48\textwidth}
         \centering
         \includegraphics[width=\textwidth]{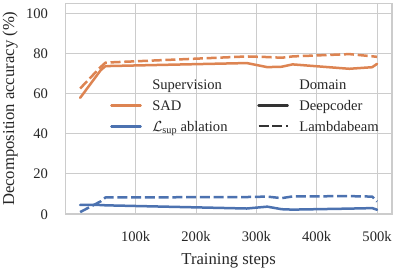}
         \caption{Training decomposition accuracy}
         \label{fig:lsup_decacc}
     \end{subfigure}
     \hfill
     \begin{subfigure}{0.48\textwidth}
         \centering
         \includegraphics[width=\textwidth]{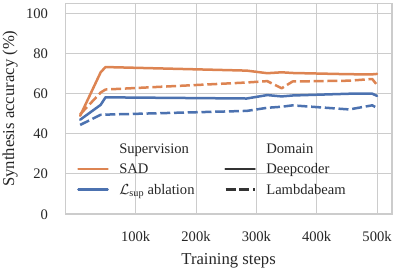}
         \caption{Training synthesis accuracy}
         \label{fig:lsup_synthacc}
     \end{subfigure}
     \caption{Training decomposition and synthesis accuracy of \ac{sad} with and without $\mathcal{L}_{\text{sup}}$ in the list domains. Without $\mathcal{L}_{\text{sup}}$, decomposition accuracy drops to 2--9\%, while synthesis accuracy on the training tasks still rises to a lower level than with $\mathcal{L}_{\text{sup}}$.}
     \label{fig:plsup_acc}
\end{figure}

To measure the contribution of $\mathcal{L}_{\text{sup}}$, we train the decomposer of \ac{sad} without it, with the objective $\mathcal{L} = \mathcal{L}_{\text{RL}} - \lambda H(\pi_\theta)$ and $\lambda = 0.01$.
With $\mathcal{L}_{\text{sup}}$, the training decomposition accuracy reaches 75--80\% within the first 50k steps and stays there (Figure~\ref{fig:lsup_decacc}).
Without it, decomposition accuracy drops to 2--9\% early in training and does not recover.
The reward still improves during this collapse: the training synthesis accuracy rises to 58--60\% on Deepcoder and 53--54\% on Lambdabeam, against 70--73\% and 65--67\% with $\mathcal{L}_{\text{sup}}$ (Figure~\ref{fig:lsup_synthacc}).
On the test tasks, however, the ablated model, which we evaluated end to end for a single seed, solves fewer than 20\% of the tasks.

This shows that without $\mathcal{L}_{\text{sup}}$, the decomposer finds degenerate subgoals from which the synthesizer still recovers part of the \ac{gt} subprograms on the training tasks, and that this does not transfer to new tasks.
It indicates that solver feedback is useful only within a range of subgoals that resemble the subgoals on which the synthesizer was trained.
$\mathcal{L}_{\text{sup}}$ keeps the decomposer within this range, and $\mathcal{L}_{\text{RL}}$ selects within it the subgoals that the synthesizer solves more reliably.
Together with the accuracy paradox (Section~\ref{sec:accuarcy_paradox}), this shows that neither signal suffices alone: with $\mathcal{L}_{\text{sup}}$ only, the decomposer reproduces \ac{gt} subgoals that the synthesizer does not always act on, and with $\mathcal{L}_{\text{RL}}$ only, it leaves the range in which the feedback of the synthesizer is informative.

\subsection{Beam Oracle}
\label{app:beamoracles}
\begin{figure}[t]
     \centering
     \begin{subfigure}[b]{0.48\textwidth}
         \centering
         \includegraphics[width=\textwidth]{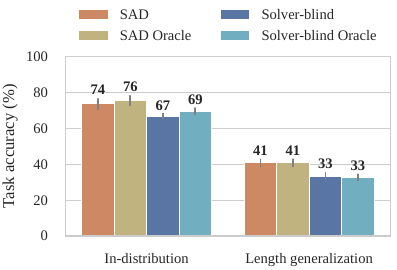}
         \caption{Deepcoder}
         \label{fig:beamoracle_dc}
     \end{subfigure}
     \begin{subfigure}[b]{0.48\textwidth}
         \centering
         \includegraphics[width=\textwidth]{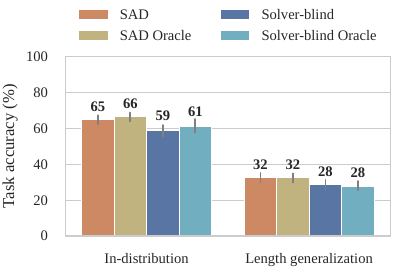}
         \caption{Lambdabeam}
         \label{fig:beamoracle_lb}
     \end{subfigure}
     \caption{Task accuracy with and without the beam oracle, which selects the \ac{gt} decomposition whenever it appears in the beam. The oracle adds less than 2 percentage points for both methods, and the solver-blind baseline with the oracle stays below \ac{sad} without it.}
     \label{fig:beamoracle}
\end{figure}

The beam oracle tests whether the gap between \ac{sad} and the solver-blind baseline would close if the decomposers ranked the \ac{gt} subgoals higher.
It selects the \ac{gt} decomposition whenever it appears in the beam and otherwise keeps the original ranking, so it separates whether a decomposer generates the \ac{gt} decomposition from how it ranks it.
For both methods and both list domains, the oracle improves task accuracy by less than 2 percentage points (Figure~\ref{fig:beamoracle}), and with the oracle, the solver-blind baseline still stays below \ac{sad} without it.
This shows that ranking the \ac{gt} subgoals higher does not close the gap.
The gap therefore lies in the decompositions that each decomposer generates, and \ac{sad} likely generates decompositions that are rare in the beam of the solver-blind decomposer.
The oracle covers only this one kind of reranking: since the \ac{gt} decomposition is not always the best one for the synthesizer (Section~\ref{sec:gtoracle}), a reranking that selects any successful candidate could give different results.

\subsection{Structure of \ac{sad}-only Solutions}
\label{app:sadonly}
On Lambdabeam, our program-parsing step fails on about 200 of the 2 × 1000 test tasks per seed due to an unsupported serialization of lambda arguments.
Whether a task is excluded depends only on the task, so \ac{sad} and the \ac{gt} oracle are compared on an identical task set.
The excluded tasks are not systematically easier or harder: on the remaining tasks, \ac{sad} solves 48.7\% and the oracle 70.4\%, against 48.7\% and 70.5\% on the full test sets.
Task accuracies in Figures~\ref{fig:perf_lb} and Table \ref{tab:pvalues} are computed on all tasks and are unaffected.

We analyze the tasks that \ac{sad} solves and the \ac{gt} oracle does not (Section~\ref{sec:gtoracle}) and compare the solutions of \ac{sad} with the \ac{gt} programs of the same tasks in overlap, length, and primitive usage.
Fewer than 2\% of these solutions match the \ac{gt} program on Deepcoder, and none does on Lambdabeam.
On Deepcoder, 47.5\% of the solutions have the same length as the \ac{gt} program, 28.2\% are longer, and 24.4\% are shorter; on Lambdabeam, 56.8\% are shorter and 17.1\% are longer.
This shows that \ac{sad} solves these tasks with different programs, of similar length on Deepcoder and mostly shorter on Lambdabeam.
The shorter programs on Lambdabeam possibly reflect its larger set of candidate subprograms per step, which gives more room for paths that need fewer steps than the \ac{gt} program.
\begin{figure}[!b]
     \centering
     \begin{subfigure}{0.48\textwidth}
         \centering
         \includegraphics[width=\textwidth]{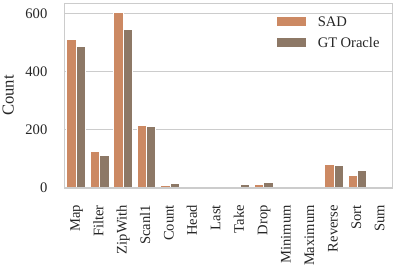}
         \caption{Deepcoder}
         \label{fig:dsl_usage_dc}
     \end{subfigure}
     \begin{subfigure}{0.48\textwidth}
         \centering
         \includegraphics[width=\textwidth]{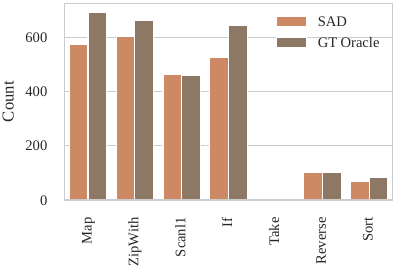}
         \caption{Lambdabeam}
         \label{fig:dsl_usage_lb}
     \end{subfigure}
     \caption{\ac{dsl} primitive usage in the solutions of \ac{sad} on tasks that the \ac{gt} oracle does not solve, compared with the \ac{gt} programs of the same tasks.}
     \label{fig:dslusage}
\end{figure}

The primitive usage shifts accordingly (Figure~\ref{fig:dslusage}).
On Deepcoder, the solutions of \ac{sad} use \texttt{ZipWith}, \texttt{Map}, and \texttt{Filter} more often and \texttt{Sort}, \texttt{Drop}, \texttt{Take}, \texttt{Count}, \texttt{Sum}, and \texttt{Head} less often than the \ac{gt} programs.
On Lambdabeam, they use almost all operations less often, except \texttt{Reverse} and \texttt{Scanl1}; since these solutions are mostly shorter, part of this decrease likely follows from their length.
Together with the low overlap, this indicates that \ac{sad} reaches the target through other primitives than the \ac{gt} programs, although its reward contains no term for deviating from \ac{gt}.

\subsection{Why a Reward Against the \ac{gt} Program Does Not Pull Toward \ac{gt} Subgoals}
\label{app:gtreward}
The reward $R(O) = -\mathrm{CE}(\mathrm{Synth}(I^{(t)}, O), p^*_t)$ contains the \ac{gt} subprogram $p^*_t$, so $\mathcal{L}_{\text{RL}}$ could in principle push the decomposer back toward the \ac{gt} subgoals.
To see why it does not, it helps to separate the \ac{gt} \emph{program}, which the reward contains, from the \ac{gt} \emph{decomposition}, which it does not.
The aim of \ac{sad} is a decomposer that proposes subgoals the synthesizer can solve, also where they leave the \ac{gt} trajectory.
A subgoal that differs from the \ac{gt} subgoal can still lead the synthesizer to $p^*_t$, and the reward favors it whenever the synthesizer recovers $p^*_t$ from it more easily.

\textbf{Why the reward uses the \ac{gt} program} \quad
Training needs a single verifiable target for the prediction of the synthesizer, and the \ac{gt} program provides it by fixing the correct semantics.
Other correct programs may exist, but finding them during training would require deciding semantic equivalence between arbitrary programs, which is computationally expensive in this setting.
The \ac{gt} program is therefore the practical target.

\textbf{What the advantage compares} \quad
The reward measures how well the frozen synthesizer recovers the correct subprogram from a proposed subgoal: $p^*_t$ fixes \emph{what} to synthesize, and the subgoal determines \emph{how} easy this is.
\ac{scst} optimizes the advantage $A = R(O^s) - R(O^g)$ and not the absolute reward.
If the sampled and the greedy subgoal lead the synthesizer to the same program, the shared target does not separate them, and $A$ only compares how confidently the synthesizer generates this program from each subgoal.
If they lead to different programs, $A$ also favors the subgoal under which the synthesizer is closer to $p^*_t$.
The reward thus prefers the semantics of the \ac{gt} program, but not the \ac{gt} path to it.

\textbf{Evidence} \quad
If $p^*_t$ pulled the decomposer toward the \ac{gt} subgoals, decomposition accuracy would rise.
Instead, \ac{sad} reaches a lower decomposition accuracy than the solver-blind baseline in every setting, together with a higher synthesis and task accuracy (Table~\ref{tab:accuracy_paradox}).
With $\mathcal{L}_{\text{RL}}$ alone, decomposition accuracy drops to 2--9\% and does not approach the \ac{gt} subgoals (Appendix~\ref{app:lsupablation}).
Finally, on the tasks that only \ac{sad} solves, its programs rarely match the \ac{gt} program and use different primitives (Appendix~\ref{app:sadonly}).
This shows that the preference of the synthesizer for subgoals it can solve dominates the training signal.
That \ac{sad} finds programs other than the \ac{gt} program is a result of this preference; using such programs as additional supervision targets is left to future work.

\setcounter{figure}{0}
\setcounter{table}{0}
\section{Details of the \ac{llm} Experiments}
\label{app:llm}
This appendix gives the setup and results of the off-the-shelf \ac{llm} baselines and the training details of \ac{sad} with Llama-3.1-8B in Section~\ref{sec:llm}.
All \acp{llm} decode with temperature 0 and generate one sample per prompt, and each count below comes from a single run.

\subsection{Off-the-Shelf \ac{llm} Baselines}
\label{app:llm_baselines}
We evaluate three open-weight \acp{llm} without fine-tuning: Llama-3.1-8B~\citep{grattafiori2024llama}, Qwen2.5-14B-Instruct~\citep{qwen2025qwen25}, and Qwen2.5-Coder-14B-Instruct~\citep{hui2024qwen25coder}.
Each model is evaluated with direct synthesis and with decomposition, where the same model predicts each subgoal and then synthesizes the subprogram for the induced subtask in the Exedec loop with at most 5 steps.
As in Section~\ref{sec:llm}, we use the 200-task Deepcoder \ac{llm} benchmark of Exedec~\citep{shi2023exedec}, which is smaller and easier than the 1000-task benchmark of the Transformer experiments, and evaluate in-distribution only.
We use the prompts of Exedec, which contain the \ac{io} specification, the list of \ac{dsl} primitives, and four few-shot demonstrations.
Programs are written in the Pythonic Deepcoder \ac{dsl} of Exedec, in which each primitive except \texttt{Scanl1} becomes an equivalent Python one-liner; for example, \texttt{x0 = INPUT | x1 = Map (*2) x0} becomes
\begin{center}
\begin{tabular}{l}
\texttt{def program(x0):} \\
\texttt{\ \ \ \ x1 = [x * 2 for x in x0]} \\
\texttt{\ \ \ \ return x1}
\end{tabular}
\end{center}

Table~\ref{tab:llm_baselines} reports the results.
No model solves more than 10\% of the tasks.
Decomposition increases the number of solved tasks for Llama-3.1-8B from 5 to 15 and for Qwen2.5-Coder-14B-Instruct from 15 to 20, and decreases it for Qwen2.5-14B-Instruct from 14 to 12.
This indicates that without fine-tuning on the \ac{dsl}, neither direct synthesis nor decomposition reaches a useful accuracy, which is why the experiment in Section~\ref{sec:llm} starts from fine-tuned adapters.
\begin{table}[t]
    \centering
    \caption{Off-the-shelf \acp{llm} on the 200-task Deepcoder \ac{llm} benchmark (in-distribution). Tasks solved out of 200.}
    \label{tab:llm_baselines}
    \small
    \setlength{\tabcolsep}{4pt}
    \begin{tabular}{lccc}
    \toprule
    \textbf{Method} & \textbf{Llama-3.1-8B} & \textbf{Qwen2.5-14B-Instruct} & \textbf{Qwen2.5-Coder-14B-Instruct} \\
    \midrule
    Direct synthesis & 5 (2.5\%) & 14 (7.0\%) & 15 (7.5\%) \\
    With decomposition & 15 (7.5\%) & 12 (6.0\%) & 20 (10.0\%) \\
    \bottomrule
    \end{tabular}
\end{table}

\subsection{\ac{sad} with Llama-3.1-8B}
\label{app:llm_sad}
Decomposer and synthesizer are two LoRA adapters~\citep{hu2022lora} on Llama-3.1-8B~\citep{grattafiori2024llama}, with rank 32, scaling factor $\alpha = 64$, and dropout 0.05 on all attention and MLP projections (\texttt{q}, \texttt{k}, \texttt{v}, \texttt{o}, \texttt{gate}, \texttt{up}, \texttt{down}).
We train them with AdamW at learning rate $10^{-4}$, a cosine schedule with 100 warm-up steps, no weight decay, gradient clipping at 1.0, and an effective batch size of 16 (4 per device with 4 gradient-accumulation steps).
Both adapters are first fine-tuned on Deepcoder for 1000 steps, after which the synthesizer is frozen.
From the same decomposer checkpoint, we then compare three conditions:
\begin{itemize}
    \item \textbf{Starting checkpoint}: the decomposer after the 1000 steps, without further training.
    \item \textbf{\Ac{sft}}: 250 further steps with $\mathcal{L}_{\text{sup}}$ only, which is the solver-blind control at equal budget.
    \item \textbf{\ac{sad}}: 250 further steps with the full objective of Section~\ref{sec:sad_training}, in which the frozen synthesizer provides the reward, the greedy subgoal serves as the \ac{scst} baseline, and $\mathcal{L}_{\text{sup}}$ and the entropy term with $\lambda = 0.01$ are kept.
\end{itemize}
The fine-tuned models use the same Pythonic \ac{dsl} and prompt as the off-the-shelf baselines, including the four few-shot demonstrations.
We evaluate on the 200 tasks of the in-distribution split with at most 5 decomposition steps and a single training seed; Table~\ref{tab:llm_sad} in Section~\ref{sec:llm_results} reports the results.

\setcounter{figure}{0}
\setcounter{table}{0}
\section{Zero-Shot Transfer to the Rule et al.\ Benchmark}
\label{app:rule}
The list-function benchmark of \citet{rule2024symbolic}, which is part of BIG-bench~\citep{srivastava2023beyond}, consists of 250 tasks that map integer lists to integer lists and were designed to study how humans learn rules, not generated from a synthesis \ac{dsl}.
This appendix describes how we select the tasks for the zero-shot transfer in Section~\ref{sec:results} and reports the full result.

\paragraph{Setup.}
We use the Lambdabeam-trained \ac{sad} and solver-blind models without retraining.
Each task of the benchmark contains several \ac{io} examples, of which we always use the first four.
As in the Lambdabeam tasks, the first three examples form the specification, and the fourth is the held-out example of the leave-one-out protocol.
To select the tasks that the Lambdabeam \ac{dsl} can express, we run enumerative search over Lambdabeam programs on the three specification examples, with the same settings as during inference on Lambdabeam, so that programs can have up to $T = 10$ steps, and keep a task if the search finds a program.
Several tasks of the benchmark are solved by the same program on different inputs; we keep only one of them, so that each selected task requires a different program.
This leaves 129 of the 250 tasks.
Metric, beam size, and step limit are the same as for Lambdabeam (Section~\ref{sec:eval}), and results are mean $\pm$ standard deviation over 5 seeds.

\paragraph{Results.}
\ac{sad} solves $42.25\% \pm 7.13\%$ of the tasks, against $32.95\% \pm 3.49\%$ for the solver-blind baseline, a gain of $9.3$ percentage points or $28.2\%$ relative.
This shows that the advantage of \ac{sad} holds on tasks from an independent source that neither model was trained on, and it indicates that the subgoals learned against the synthesizer are not specific to the task generator of Lambdabeam.
The larger standard deviation of \ac{sad} likely reflects the small number of tasks and the shift between the task distributions.

\end{document}